\documentclass[acmtog]{acmart}
\pdfoutput=1
\usepackage{microtype}

\usepackage{amsthm, bm}
\usepackage{amsthm,amssymb}
\usepackage[linesnumbered,ruled]{algorithm2e}
\usepackage[english]{babel}
\usepackage{wrapfig}
\usepackage{subfigure} 
\usepackage{soul}
\usepackage{enumitem}
\usepackage{amsfonts}
\usepackage{booktabs}
\usepackage{multirow}
\usepackage{siunitx}

\setcitestyle{square}

\usepackage[ruled]{algorithm2e} 

\SetAlFnt{\small}
\SetAlCapFnt{\small}
\SetAlCapNameFnt{\small}
\SetAlCapHSkip{0pt}
\IncMargin{-\parindent}

\acmJournal{TOG}

\setcopyright{usgovmixed}

\received{July 2017}

\newcommand\relphantom[1]{\mathrel{\phantom{#1}}}

\newcommand{\figspace}[0]{\vspace{-3.5mm}}

\newcommand{\sLNM}{}
\newcommand{\tLNM}{}

\newcommand{\figref}[1]{Fig.~\ref{#1}}

\newcommand{\eqnref}[1]{Eq.~\eqref{#1}}
\newcommand{\secref}[1]{\S\ref{#1}}

\newcommand{\bb}[1]{\mathbb{#1}}
\newcommand{\br}[0]{\bm{r}}

\newcommand{\mdots}{...}

\begin{document}
\title{FontCode: Embedding Information in Text Documents using Glyph Perturbation}

\author{Chang Xiao}
\author{Cheng Zhang}
\author{Changxi Zheng}
\affiliation{%
  \institution{Columbia University}
  \department{School of Engineering and Applied Science}
}

\renewcommand\shortauthors{Xiao, C. et al}

\begin{abstract}
    We introduce \emph{FontCode}, an information embedding technique for text
    documents. Provided a text document with specific fonts, our method embeds
    user-specified information in the text by perturbing the glyphs of text
    characters while preserving the text content.  We devise an algorithm to
    choose unobtrusive yet machine-recognizable glyph perturbations, 
    leveraging a recently developed generative model that alters the glyphs of
    each character continuously on a font manifold.  
    We then introduce an algorithm that embeds a user-provided message in the text document 
    and produces an encoded document whose appearance is minimally perturbed from the original document. 
    We also present a glyph recognition method that recovers the
    embedded information from an encoded document stored as a vector
    graphic or pixel image, or even on a printed paper.  
    In addition, we introduce a new error-correction coding scheme that
    rectifies a certain number of recognition errors.  Lastly, we demonstrate
    that our technique enables a wide array of applications, 
    using it as a text document metadata holder, an unobtrusive
    optical barcode, a cryptographic message embedding scheme, and a text document
    signature.
\end{abstract}

%
%


%
%


%

\begin{teaserfigure}
   \vspace{-1mm}
   \centering
   \includegraphics[width=0.95\textwidth]{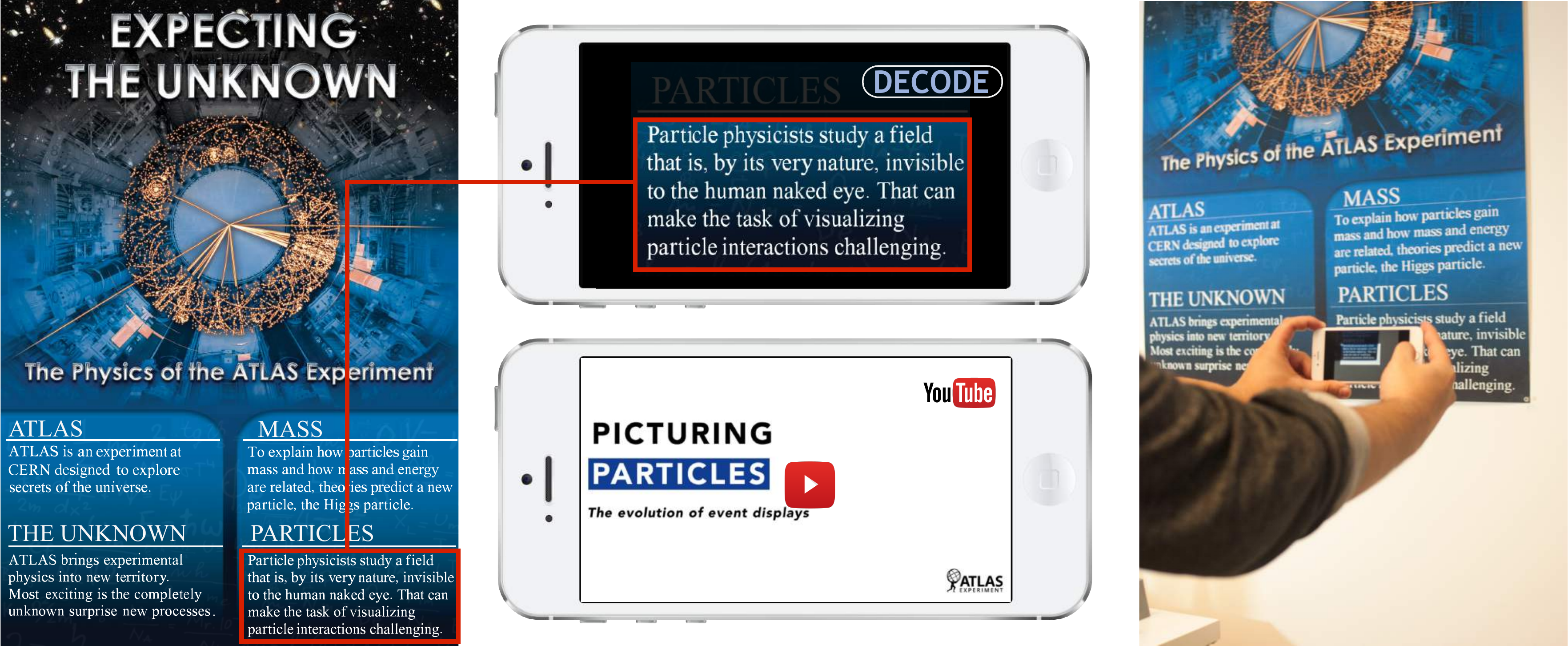}
   \vspace{-3mm}
   \caption{ {\bf Augmented poster.} 
   Among many applications enabled by our FontCode method, here we create a poster embedded
   with an unobtrusive optical barcode (left). 
   The poster uses text fonts that look almost identical from the standard Times New Roman, 
   and has no traditional black-and-white barcode pattern.
   But our smartphone application allows the user to take 
   a snapshot (right) and decode the hidden message, in this case, a Youtube link (middle).
   }\label{fig:teaser}
\end{teaserfigure}

\maketitle

\section{Introduction}
Information embedding, the technique of embedding a message into host data, has numerous applications:
Digital photographs have metadata embedded to record such information as capture date, exposure time, focal
length, and camera's GPS location.
Watermarks embedded in images, videos, and audios
have been one of the most important means in digital production to claim copyright against piracies~\cite{bloom1999copy}.
And indeed, the idea of embedding information in light signals has grown into an emerging field of visual light 
communication (e.g., see~\cite{Jo2016DDC}).

In all these areas, information embedding techniques meet two desiderata:
(i) the host medium is minimally perturbed, implying that the embedded message
must be minimally intrusive; and 
(ii) the embedded message can be robustly recovered by the intended decoder even in the presence of
some decoding errors.


Remaining reclusive is the information embedding technique for text documents, in both digital and physical form. 
While explored by many previous works on digital text steganography,
information embedding for text documents is considered more challenging 
to meet the aforementioned desiderata
than its counterparts for images, videos, and audios~\cite{agarwal2013text}.
This is because the ``pixel'' of a text document is individual letters, which, 
unlike an image pixel, cannot be changed into other letters without causing noticeable differences.
Consequently, existing techniques have limited information capacity or
work only for specific digital file formats (such as PDF \emph{or} Microsoft Word).

We propose \emph{FontCode}, a new information embedding technique for text
documents.  Instead of changing text letters into different ones, we alter the
\emph{glyphs} (i.e., the particular shape designs) of their fonts to encode 
information, leveraging the
recently developed concept of font manifold~\cite{Campbell2014LMF} in computer
graphics. Thereby, the readability of the original document is fully retained.
We carefully choose the glyph perturbation such that it has a minimal effect
on the typeface appearance of the text document, while ensuring that glyph perturbation
can be recognized through Convolutional Neural Networks (CNNs).
To recover the embedded information, 
we develop a decoding algorithm that recovers the information from an input encoded 
document---whether it is represented as a vector graphics file (such as a PDF) or 
a rasterized pixel image (such as a photograph).

Exploiting the features specific to our message embedding and retrieval problems,
we further devise an error-correction coding scheme that is able to fully recover
embedded information up to a certain number of recognition errors,
making a smartphone into a robust FontCode reader (see \figref{fig:teaser}).

\paragraph{Applications}
As a result, FontCode is not only an information embedding technique for text documents
but also an unobtrusive tagging mechanism, finding a wide array of applications.
We demonstrate four of them.
(i) It serves as a metadata holder in a text document, which can be freely converted to different file formats or 
printed on paper without loss of the metadata---across various digital and physical forms, the metadata is always preserved.
(ii) It enables to embed in a text unobtrusive optical codes, ones that can replace
optical barcodes (such as QR codes) in artistic designs such as posters and flyers to minimize visual distraction caused by the barcodes.
(iii) By construction, it offers a basic cryptographic scheme that not only embeds but also encrypts messages, without 
resorting to any additional cryptosystem.
And (iv) it offers a new text signature mechanism, one that allows to verify document authentication and integrity, regardless
of its digital format or physical form.

\paragraph{Technical contributions}
We propose an algorithm to construct a glyph
codebook, a lookup table that maps a message into perturbed
glyphs and ensures the perturbation is unobtrusive to
our eyes. We then devise a recognition method that recovers the embedded
message from an encoded document.
Both the codebook construction and glyph recognition leverage CNNs.
Additionally, we
propose an error-correction coding scheme that rectifies recognition errors due
to factors such as camera distortion. Built on a 1700-year old number theorem,
the Chinese Remainder Theorem, and a probabilistic decoding model, our
coding scheme is able to correct more errors than what
block codes based on Hamming distance can correct, 
outperforming their theoretical error-correction upper bound.


\section{Related Work}
We begin by clarifying a few typographic terminologies~\cite{Campbell2014LMF}: 
the \emph{typeface} of a character refers to a set of \emph{fonts} each
composed of \emph{glyphs} that represent the specific design and features of the
character. With this terminology, our method embeds messages by perturbing 
the glyphs of the fonts of text letters.


\paragraph{Font manipulation}
While our method perturbs glyphs using the generative model
by Campbell and Kautz~\shortcite{Campbell2014LMF}, other methods 
create fonts and glyphs with either computer-aided tools~\cite{rugglcs1983letterform}
or automatic generation.
The early system by~\cite{knuth1986metafont} creates parametric fonts 
and was used to create most of the Computer Modern typeface family.
Later Shamir and Rappoport~\shortcite{shamir1998feature} proposed a system 
that generate fonts using
high-level parametric features and constraints to adjust glyphs.
This idea was extended to parameterize glyph shape components~\cite{hu2001parameterizable}.
Other approaches generate fonts by deriving from examples 
and templates~\cite{Lau:2009:LEP,suveeranont2010example}. 
More recent works have focused on structuring fonts based on
similarity~\cite{loviscach2010universe} or crowdsourced attributes~\cite{ODonovan2014}.

\paragraph{Font recognition}
Automatic font recognition from a photo or image has been studied~\cite{aviles2005high,jung1999multifont,ramanathan2009novel}.
These methods identify fonts by extracting statistical and/or typographical features of the document.
Recently in~\cite{chen2014large}, the authors proposed a scalable solution leveraging supervised learning.
Then, Wang et al.~\shortcite{Wang:2015:DIY:2733373.2806219} improved font recognition 
using Convolutional Neural Networks. Their algorithm can run without resorting to 
character segmentation and optical character recognition methods.
In our system, we use existing algorithms to recognize text fonts of the input
document, but further devise an algorithm to recognize glyph perturbation for recovering
the embedded information. Unlike existing font recognition methods that identify fonts from a text 
of many letters, our algorithm aims to identify glyph perturbation for individual letters.




\paragraph{Text steganography}

Our work is related to digital steganography (such as digital watermarks for copyright protection), 
which has been studied for decades, mostly focusing on videos, images, and audios. 
However, digital text steganography
is much more challenging~\cite{agarwal2013text},
and thus much less developed.
We categorize existing methods based on their features (see Table~\ref{tb:compare}):



Methods based on cover text generation (\textbf{CTG}) hide a secrete message by
generating an \emph{ad-hoc} cover text which looks lexically and syntactically
convincing~\cite{wayner1992mimic,wayner2009disappearing}.
However, this type of steganography is unable to embed messages in existing text documents.
Thus, they fail to meet the attribute that we call cover text preservation (\textbf{CP}),
and have limited applications.

The second type of methods exploits format-specific features (\textbf{FSF}).
For example, specifically for Microsoft Word document, Bhaya et al.~\shortcite{bhaya2013text}
assign each text character a different but visually similar font available in Word 
to conceal messages. 
Others hide messages by changing the character scale and color or adding underline styles 
in a document~\cite{stojanov2014new,pandatext}, although those changes are generally noticeable.
More recent methods exploit special ASCII codes and Unicodes that are displayed as an
empty space in a PDF viewer~\cite{chaudhary2016text,rizzo2016content}.
These methods are not format independent (\textbf{FI}); they are bounded to a specific file format (such 
as Word or PDF) and text viewer. The concealed messages would be lost if the document was converted 
in a different format or even opened with a different version of the same viewer.
If also fails to preserve the concealed message when the document is printed on paper (\textbf{PP}) 
and photographed later.


More relevant to our work is the family of methods that embed messages  
via what we call structural perturbations (\textbf{SP}).
Line-shift methods~\cite{alattar2004watermarking,brassil1995electronic} hide 
information by perturbing the space between text lines: reducing or increasing
the space represents a 0 or 1 bit.
Similarly, word-shift methods~\cite{kim2003text,brassil1995electronic}
perturb the spaces between words.
Others perturb the shape of specific characters, such as raising or dropping the
positions of the dots above ``i'' and ``j''~\cite{brassil1995electronic}.
These methods cannot support fine granularity coding (\textbf{FGC}), in the sense that
they encode 1 bit in every line break, word break or special character that appears
sparsely. Our method provides fine granularity coding by embedding information in 
individual letters, and thereby has a much larger information capacity.
In addition, all these methods demonstrate retrieval of hidden messages from \emph{digital document files only}. 
It is unclear to what extent they can decode from real photos of text.

\begin{table} 
  \begin{tabular}{cccccc}
    \toprule
    \centering
    \multirow{2}{*}{Category} &
    \multirow{2}{*}{Previous work} &
    \multicolumn{4}{c}{Attribute} \\
    & {} & {CP} & {FGC} & {FI} & {PP} \\
    \midrule
    \multirow{2}{*}{CTG} & \multirow{2}{3cm}{~\cite{wayner2009disappearing,agarwal2013text}} & & \multirow{2}{*}{\checkmark} &  \multirow{2}{*}{\checkmark}  &  \multirow{2}{*}{\checkmark} \\  \\ \hline
    \multirow{5}{*}{SP} & \multirow{5}{3cm}{ ~\cite{brassil1995electronic,kim2003text,alattar2004watermarking,gutub2007novel}} & \multirow{5}{*}{\checkmark} & & 
    \multirow{5}{*}{\checkmark} & \multirow{5}{*}{\checkmark} \\ \\  \\ \\ \\ \hline
    \multirow{4}{*}{FSF} & \multirow{2}{3cm}{~\cite{pandatext,bhaya2013text,chaudhary2016text,rizzo2016content}} & \multirow{4}{*}{\checkmark} & \multirow{4}{*}{\checkmark} & & 
    \\ \\ \\ \\ \hline
    \textbf{Our work} & & \checkmark & \checkmark & \checkmark & \checkmark \\ \hline
  \end{tabular}
  \caption{
      {\bf A summary} of related text steganographic methods.
      CP indicates cover text preservation; FGC indicates fine
  granularity coding; FI indicates format-independent; PP indicates printed paper.
  \label{tb:compare}}
  \vspace{-8mm}
\end{table}

Table~\ref{tb:compare} summarizes these related work and their main attributes.
Like most of these work, our method aims to perturb the appearance of the text 
in an unobtrusive, albeit not fully imperceptible, way.
But our method is advantageous by providing more desired attributes.

\section{Overview}
Our FontCode system embeds in a text document any type of information as a bit string.
For example, an arbitrary text message can be coded into a bit string using
the standard ASCII code or Unicode\footnote{See the character set coding standards by the Internet Assigned Numbers Authority (\url{http://www.iana.org/assignments/character-sets/character-sets.xhtml})}.
We refer to such a bit string as a \emph{plain message}.

In a text document, the basic elements of embedding a plain message are the
\emph{letters}, appearing in a particular font.  Our idea is to perturb the
glyph of each letter to embed a plain message. To this end, we leverage the
concept of \emph{font manifold} proposed by Campbell and
Kautz~\shortcite{Campbell2014LMF}. Taking as input a collection of existing
font files, their method creates a low-dimensional font manifold for every
character---including both alphabets and digits---such that every location on
this manifold generates a particular glyph of that character. 
This novel generative model is precomputed once for each character.
Then, it allows us to alter the glyph of each text letter in a subtle yet
systematic way, and thereby embed messages.

In this paper, 
when there is no confusion,
we refer to a location $\bm{u}$ on a font manifold and its resulting glyph
interchangeably. 

\begin{figure}[t]
  \centering
  \figspace
  \includegraphics[width=1.0\columnwidth]{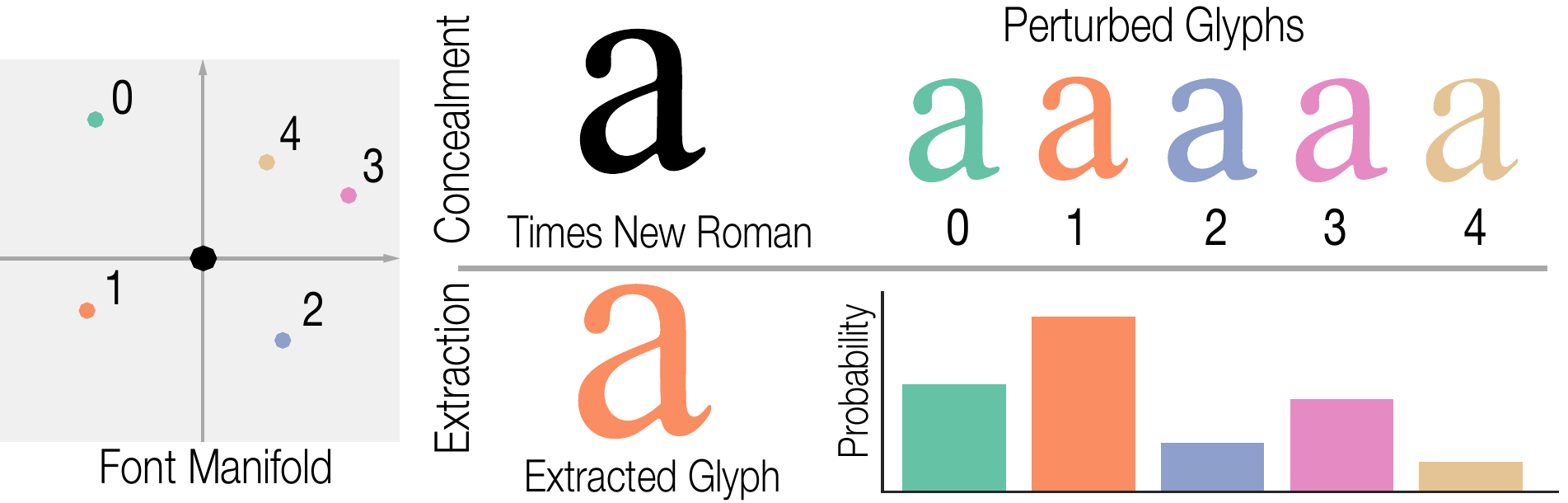}
  \vspace{-6mm}
  \caption{ {\bf Embedding and extraction.} 
  Here we sample 5 points around the Times New Roman on the manifold (left), 
  generating the perturbed glyphs to embed integers (top-right).
  We embed ``1'' in letter ``a''  using the second glyph (in orange) in the perturbation list.
  In the retrieval step, we evaluate a probability value (inverse of distance) by our CNNs (bottom-right), 
  and extract the integer whose glyph results in the highest probability.
  }\label{fig:code}
  \vspace{-5mm}
\end{figure}

\begin{figure*}[t]
  \centering
  \includegraphics[width=0.98\textwidth]{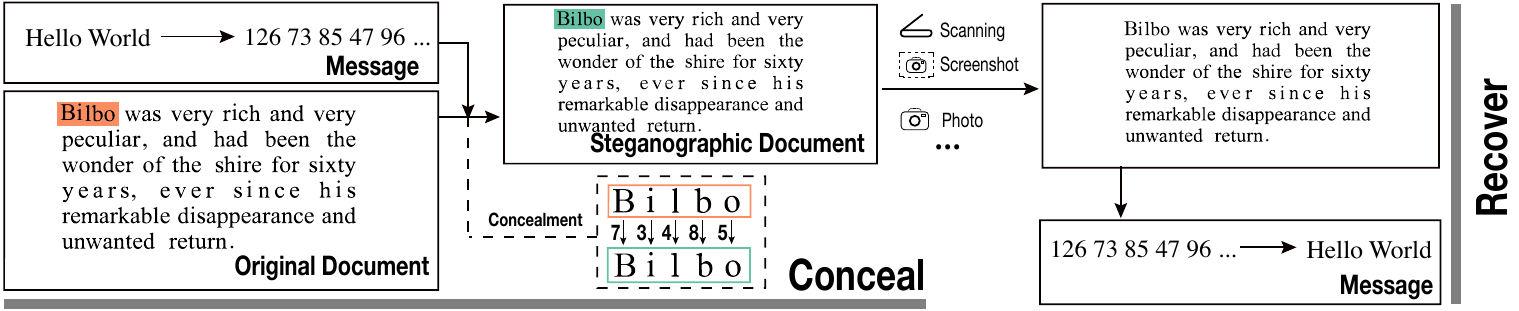}
  \vspace{-2mm}
  \caption{ {\bf Overview.} (left) Our embedding method takes an input message and a text document. It encodes the message 
  into a series of integers and divides the letters into blocks. The integers are assigned to each block and embedded in individual letters. 
  (right) To recover the message, we extract integers by recognizing the glyphs of individual letters.
  Then, the integers are decoded into the original plain message.
  } \label{fig:overview}
  \figspace
\end{figure*}

\subsection{Message Embedding}\label{subsec:encode}
Our message embedding method consists of two steps, (i) precomputation of a codebook
for processing all documents and (ii) runtime embedding of a plain message in
a given document.

During precomputation, we construct a codebook of perturbed 
glyphs for typically used fonts.  Consider a font such as
Times New Roman. Each character in this font corresponds to a specific
location, $\bar{\bm{u}}$, on the font manifold.  We identify a set of locations
on the manifold as the \emph{perturbed glyphs} of $\bar{\bm{u}}$
and denote them as $\{\bm{u}_0,\bm{u}_1,\ldots\}$ (see \figref{fig:code}). Our
goal in this step is to select the perturbed glyphs such
that their differences from the glyph $\bar{\bm{u}}$ of the original font is
almost unnoticeable to our naked eyes but recognizable to a computer algorithm
(detailed in \S\ref{sec:recog}).  The sets of perturbed 
glyphs for all characters with typically used fonts form our codebook.

At runtime, provided a text document (or a text region or paragraphs), 
we perturb the glyph of the letter in the
document to embed a given plain message.
Consider a letter in an original glyph $\bar{\bm{u}}$ in the given document.
Suppose in the precomputed codebook, this letter has $N$ perturbed
glyphs, namely $\{\bm{u}_0,\bm{u}_1,\ldots,\bm{u}_{N-1}\}$.  We
embed in the letter an integer $i$ in the range of $[0,N)$ by changing 
its glyph from $\bar{\bm{u}}$ to $\bm{u}_i$ (see \figref{fig:code}-top). 
A key algorithmic component of this step is to
determine the embedded integers for all letters such that together they
encode the plain message. 


In addition, we propose a new error-correcting coding scheme to encode the plain
message. This coding scheme adds certain redundancy (i.e., some extra data) to
the coded message, which, at decoding time, can be used to check for
consistency of the coded message and recover it from errors (see \secref{sec:coding}).

\subsection{Message Retrieval}\label{subsec:decode}
To retrieve information from a coded text document, 
the first step is to recover an integer from each
letter.  For each letter in the document,
suppose again this letter has $N$ perturbed glyphs in the codebook, whose
manifold locations are $\{\bm{u}_0,\bm{u}_1,\ldots,\bm{u}_{N-1}\}$.  
We recognize its glyph $\bm{u}'$ in the current document as 
one of the $N$ perturbed glyphs. 
We extract an integer $i$ if $\bm{u}'$ is recognized as $\bm{u}_i$ (see Fig.~\ref{fig:code}-bottom).
Our recognition
algorithm works with not only vector graphics documents (such as those stored as
PDFs) but also rasterized documents stored as pixel images. For the latter, the recognition leverages convolutional neural networks.

The retrieved integers are then fed into our error correction coding
scheme to reconstruct the plain message.  Because of the data redundancy, even
if some glyphs are mistakenly recognized (e.g., due to poor image
quality), those errors will be rectified and we will still be able to
recover the message correctly (see \S\ref{sec:coding}).

\section{Glyph Recognition}\label{sec:recog}
We start by focusing on our basic message embedding blocks: individual letters
in a text document.  Our goal in this section is to embed an integer number in
a single letter by perturbing its glyph, and later retrieve that integer from a vector
graphics or pixel image of that letter.  In the next section, we will address
what integers to assign to the letters in order to encode a message.

As introduced in \S\ref{subsec:encode}, we embed an integer in a letter through glyph perturbation, 
by looking up a precomputed codebook.
Later, when extracting an integer from a letter, we compute a ``distance'' metric between the extracted glyph and
each perturbed glyph in $\{\bm{u}_0,\ldots,\bm{u}_{N-1}\}$ in the codebook; we obtain integer $i$ if the ``distance''
of glyph $\bm{u}_i$ is the smallest one.

Our recognition algorithm supports input documents stored as vector graphics and
pixel images.  While it is straightforward to recognize vector graphic glyphs,
pixel images pose significant challenges due to camera perspectives,
rasterization noise, blurriness, and so forth.
In this section, we first describe our algorithm that decodes integers from rasterized (pixel) glyphs. 
This algorithm leverages convolutional neural networks (CNNs), which also allow us to 
systematically construct the codebook of perturbed glyphs.
Then, we describe the details of embedding and extracting integers, as well as a
simple algorithm for recognizing vector graphic glyphs. 

%

\subsection{Pixel Image Preprocessing}\label{subsec:preprocess}
When a text document is provided as a pixel image, we use the off-the-shelf
optical character recognition (OCR) library to detect individual letters.
Our approach does not depend on any particular OCR library. In practice, we
choose to use \textsf{Tesseract}~\cite{Smith2007OTO}, one of the most popular
open source OCR engines. In addition to detecting letters, OCR also identifies
a bounding box of every letter on the pixel image.

\begin{figure*}[t]
  \centering
  \figspace
  \includegraphics[width=0.98\textwidth]{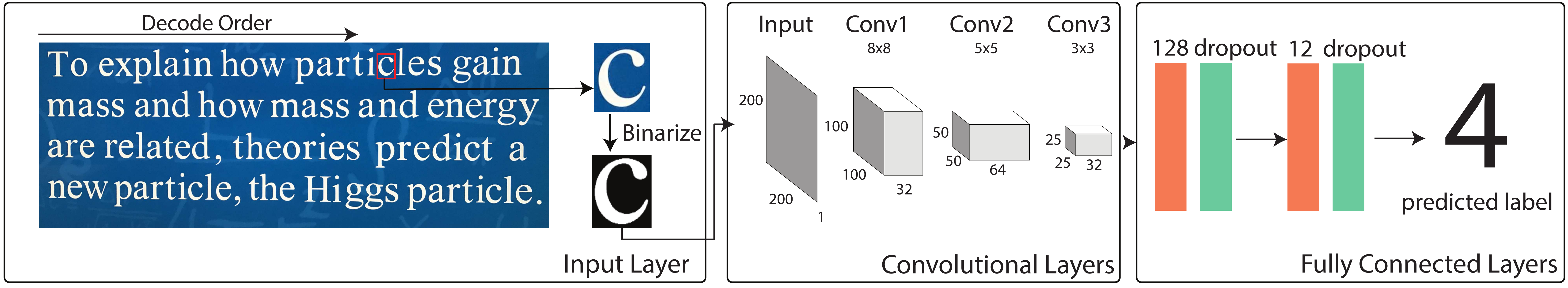}
  \vspace{-2.5mm}
   \caption{{\bf Network architecture.}
   Our CNN takes as input a 200$\times$200, black-and-white image containing a
   single letter (left). It consists 3 convolutional layers (middle) and 2 fully connected
   layers (right). It outputs a vector, in which each element is the estimated probability of recognizing
   the glyph in the input image as a perturbed glyph in the codebook. 
   }
 \label{fig:network}
 \figspace
  \vspace{1mm}
\end{figure*}

To recognize the glyph perturbation of each letter using CNNs, 
we first preprocess the image. We crop the region of each letter using its
bounding box detected by the OCR. We then binarize the image region using
the classic algorithm by Otsu~\shortcite{otsu1975threshold}. 
This step helps to eliminate the influence caused by the variations of lighting conditions and background colors.
Lastly, we resize the image region to have 200$\times$200
pixels. This 200$\times$200, black-and-white image for each letter is the
input to our CNNs (\figref{fig:network}-left).

\subsection{Network Structure}\label{subsec:architecture}
We treat glyph recognition as an image classification problem: provided an
image region of a letter which has a list of perturbed glyphs $\{\bm{u}_0,\bm{u}_1,\ldots\}$
in the codebook, our goal is to classify the input glyph of that letter as one from the list.
Therefore, we train a CNN for each letter in a particular font.

Thanks to the image preprocessing, we propose to use a simple CNN structure (as illustrated in~\figref{fig:network}),
which can be quickly trained and evaluated. The input is a 200$\times$200, black-and-white image containing a letter.
The CNN consists of three convolutional layers, each followed by a ReLU
activation layer (i.e., $f:x\in\mathbb{R}\mapsto\textrm{max}(0,x)$) and a
2$\times$2 max pooling layer. 
The kernel size of the three convolutional layers are 8$\times$8$\times$32,
5$\times$5$\times$64, and 3$\times$3$\times$32, respectively. 
The convolutional layers are connected with two fully connected (FC) layers, 
arranged in the following form (see~\figref{fig:network}):
$$
\to \textrm{FC}(128)\to\textrm{Dropout}\to\textrm{FC}(N)\to\textrm{Softmax}\to\textrm{output}.
$$
Here ``Dropout'' indicates a 0.5-dropout layer, and FC$(m)$ is an m-dimensional FC layer.
In the second FC layer, $N$ is the number of perturbed glyphs in the codebook for the letter in a certain font.
After passing through a softmax layer, the output is an $N$-dimensional vector
indicating the probabilities (or the inverse of ``distance'')
of classifying the
glyph of the input letter as one of the perturbed glyphs in the list. The glyph of the letter is recognized as $\bm{u}_i$
if its corresponding probability in the output vector is the largest one.


\subsection{Network Training}\label{subsec:training}
\paragraph{Training data}
Our CNNs will be used for recognizing text document images that are either directly synthesized or captured by digital cameras.
Correspondingly, the training data of the CNNs consist of synthetic images and real photos.
Consider a letter whose perturbed glyphs from a standard font are
$\{\bm{u}_0,\ldots,\bm{u}_{N-1}\}$. We print all the $N$ glyphs on a paper and
take 10 photos with different lighting conditions and slightly different camera angle. While taking the photos,
the camera is almost front facing the paper, mimicking the scenarios of how the
embedded information in a document would be read by a camera. 
In addition, we include synthetic images by rasterizing each glyph (whose
vector graphics path is generated using~\cite{Campbell2014LMF}) into a
200$\times$200 image.


%

\paragraph{Data augmentation}
To train the CNNs, each training iteration randomly draws a certain number
of images from the training data. We ensure that half of the images are
synthetic and the other half are from real photos. To reduce overfitting and
improve the robustness, we augment the selected images before feeding them into
the training process. Each image is processed by the following operations:
\begin{itemize}[itemsep=0pt,topsep=2pt,leftmargin=3.8mm]
  \item Adding a small Gaussian noise with zero mean and standard deviation 3.
  \item Applying a Gaussian blur whose standard deviation is uniformly distributed between 0 (no blur) and 3px.
  \item Applying a randomly-parameterized perspective transformation.
  \item Adding a constant border with a width randomly chosen between 0 and 30px. 
\end{itemize}
We note that similar data augmentations have been used in~\cite{Wang:2015:DIY:2733373.2806219,chen2014large} to enrich their training data.
Lastly, we apply the preprocessing routine described in \S\ref{subsec:preprocess},
obtaining a binary image whose pixel values are either 0 or 1.

  
  

\paragraph{Implementation details}
In practice, our CNNs are optimized using the Adam algorithm~\cite{kingma2014adam} with the parameters $\beta_1=0.9$, $\beta_2=0.999$ 
and a mini-batch size of 15.
The network was trained with $10^5$ update iterations at a
learning rate of $10^{-3}$. 
Our implementation also leverages existing libraries, Tensorflow~\cite{tensorflow2015-whitepaper} and
Keras~\cite{chollet2015keras}.


\subsection{Constructing The Glyph Codebook}\label{subsec:codebook}
The CNNs not only help to recognize glyphs at runtime, but also enable us to systematically
construct the glyph codebook, a lookup table that includes
a set of perturbed glyphs for every character in commonly used fonts 
(such as Times New Roman, Helvetica, and Calibri). 
Our codebook construction aims to satisfy three criteria:
(i) The included glyph perturbation must be perceptually similar; their differences from 
the original fonts should be hardly noticeable to our eyes.
(ii) The CNNs must be able to distinguish the perturbed glyphs reliably.
(iii) Meanwhile, we wish to include as many perturbed glyphs as possible in the
codebook for each character, in order to increase the capacity of embedding information (see \S\ref{sec:coding}).

To illustrate our algorithm, consider a single character in a given font.  We
use its location $\bar{\bm{u}}$ on the character's font manifold to refer to
the original glyph of the font.  Our construction algorithm first
finds a large set of glyph candidates that are perceptually similar to
$\bar{\bm{u}}$. This step uses crowdsourced perceptual studies 
following a standard user-study protocol. We therefore defer its details until
\S\ref{sec:perceptual}, while in this section we denote the resulting set of glyphs as $\mathcal{C}$,
which typically has hundreds of glyphs.
Next, we reduce the size of $\mathcal{C}$ by discarding glyphs that may confuse our CNNs in recognition tests.
This is an iterative process, wherein each iteration takes the following two steps (see Algorithm~\ref{alg:cftest} in Appendix):
\textbf{Confusion test.}
We randomly select $M$ pairs of glyphs in $\mathcal{C}$ ($M=$100 in practice). 
For each pair, we check if our CNN structure (introduced in
\S\ref{subsec:architecture}) can reliably distinguish the two glyphs. In this
case, we consider only the two glyphs, and thus the last FC layer (recall
\S\ref{subsec:architecture}) has two dimensions. We train this network with
only synthetic images processed by the data augmentations. We use synthetic
images generated by different augmentations to test the accuracy of the
resulting CNN. If the accuracy is less than 95\%, then we record this pair of
glyphs in a set $\mathcal{D}$. 


\textbf{Updating glyph candidate set.}
Next, we update the glyph candidates in $\mathcal{C}$ to avoid the recorded glyph pairs 
that may confuse the CNNs while retaining as many glyph
candidates as possible in $\mathcal{C}$.  To this end, we construct a graph where every
node $i$ represents a glyph in $\mathcal{C}$.  Initially, this graph is
completely connected.  Then, the edge between node $i$ and $j$ is removed if
the glyph pair $i$ $j$ is listed in $\mathcal{D}$, meaning that $i$ and
$j$ are hardly distinguishable by our CNN.  With this graph, we find the
maximum set of glyph that can be distinguished from each other. This amounts to
the classic maximum clique problem on graphs.  Although the maximum clique problem
has been proven NP-hard (see page 97 of~\cite{karp1972reducibility}), our graph size is small 
(with up to 200 nodes), and the edges are sparse.  Consequently, many existing algorithms
suffice to find the maximum clique efficiently.  In practice, we
choose to use the method of~\cite{konc2007improved}.  Lastly, the glyphs
corresponding to the maximum clique nodes form the updated glyph candidate set
$\mathcal{C}$, and we move on to the next iteration.

This iterative process stops when there is no change of $\mathcal{C}$. In our
experiments, this takes up to 5 iterations. 
Because only synthetic images are used as training data, 
this process is fully automatic and fast, but it narrows down the glyph
candidates conservatively---two glyphs in $\mathcal{C}$ might
still be hardly distinguishable in real photos.  Therefore, in a last step, we
train a CNN as described in \S\ref{subsec:training} using both synthetic images
and real photos to verify if the CNN can reliably recognize all glyphs in
$\mathcal{C}$. Here the last FC layer of CNN has a dimension of $|\mathcal{C}|$ (i.e., the size of $\mathcal{C}$). 
At this point, $|\mathcal{C}|$ is small (typically around 25). Thus, the CNN can be easily trained.
Afterwards, we evaluate the recognition accuracy of the CNN for each glyph in $\mathcal{C}$,
and discard the glyphs whose recognition accuracy is below 90\%. The remaining glyphs in $\mathcal{C}$ are
added to the codebook as the perturbed glyphs of the character.

\subsection{Embedding and Retrieving Integers}\label{subsec:embed}

\paragraph{Embedding}
Now we can embed integers into a text document 
in either vector graphic
or pixel image format. First, we extract from the input document the text content and the layout of the letters.
Our method also needs to know the original font $\bar{\bm{u}}$ of the text. This can be specified by the user or
automatically obtained from the metadata of a vector graphic document (such as a PDF). If the document is provided
as a pixel image, recent methods~\cite{chen2014large,Wang:2015:DIY:2733373.2806219} can be used to recognize the text font.

In order to embed an integer $i$ in a letter of font $\bar{\bm{u}}$, we look up its perturbed glyph list $\{\bm{u}_0,\ldots,\bm{u}_{N-1}\}$
in the precomputed codebook, and generate the letter shaped by the glyph $\bm{u}_i$~\cite{Campbell2014LMF}.
We then scale the glyph to fit it into the bounding box of the original letter (which is
detected by the OCR library for pixel images), and use it to replace the original letter of the input document.

\begin{figure}[t]
  \centering
  \includegraphics[width=0.99\columnwidth]{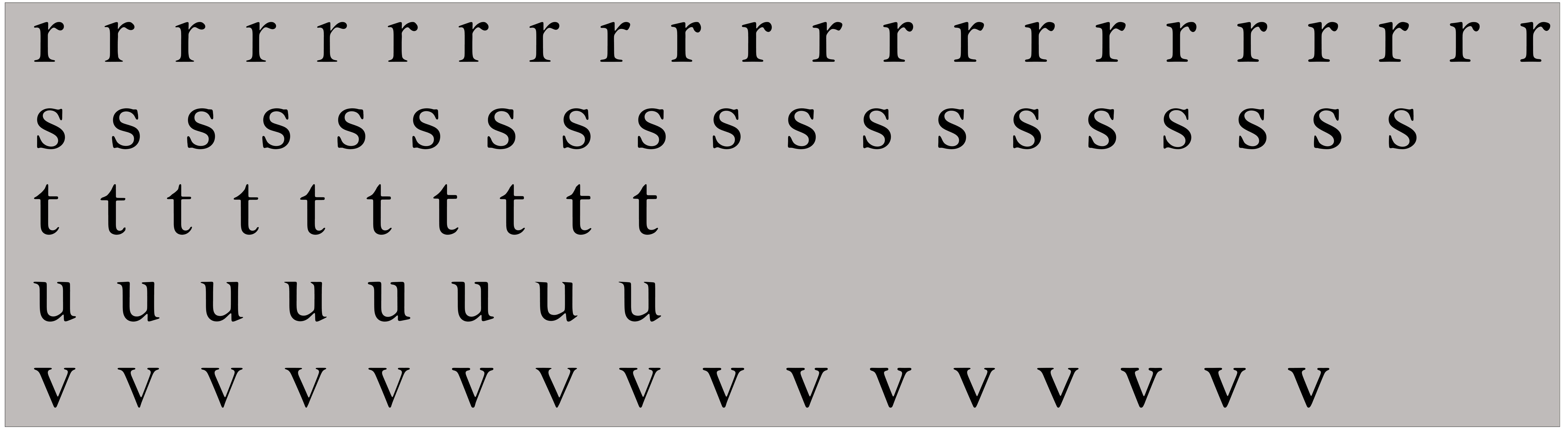} 
  \vspace{-3mm}
  \caption{ {\bf A fragment of codebook.} A complete codebook
  for all alphabetic characters in lower case for Times New Roman can be 
  found in the supplemental document.
  } \label{fig:codebook}
  \figspace
\end{figure}

\paragraph{Retrieval}
To retrieve integers from a pixel image, we extract the text content and identify regions of individual letters using 
the OCR tool. After we crop the regions of the letters, they are processed in parallel by the recognition 
algorithm: they are preprocessed as described in \S\ref{subsec:preprocess} and then fed into the CNNs. An integer $i$ is extracted
from a letter if the output vector from its CNN has the $i$-th element as the largest (recall \S\ref{subsec:architecture}).

If the input document is provided as a vector graphics image (e.g., as a PDF file), glyph recognition of 
a letter is straightforward. In our codebook, we store the outlines of the perturbed glyphs as polylines.
In the integer embedding stage, these polylines are scaled to replace the paths of the original glyphs. 
In the retrieval stage, we compute the ``distance'' between the letter's glyph polyline, denoted as $\bm{f}$, in the document and
every glyph $\bm{u}_i$ in the perturbed glyph list of the codebook:
we first scale the polylines of $\bm{u}_i$ to match the bounding box of $\bm{f}$, and then compute the $L_2$ distance
between the polyline vertices of $\bm{u}_i$ and $\bm{f}$. We recognize $\bm{f}$ as $\bm{u}_j$ (and thus extract the integer $j$)
if the distance between $\bm{f}$ and $\bm{u}_j$ is the minimum.

\section{Error Correction Coding}\label{sec:coding}
After presenting the embedding and retrieval of integer values over individual letters, 
we now focus on a plain message represented as a bit string
and describe our coding scheme
that encodes the bit string into a series of integers, which will be
in turn embedded in individual letters. 
We also introduce an error-correction decoding algorithm that decodes the bit
string from a series of integers, even when some integers are incorrect.

Algorithms presented in this section are built on the coding theory for noise channels, founded by 
Shannon's landmark work~\cite{Shannon1948}. 
We refer the reader to the textbook~\cite{lin2004error} for a comprehensive introduction.
Here, we start with a brief introduction of 
traditional error-correcting codes, to point out their fundamental differences from 
our coding problem.


\subsection{Challenges of the Coding Problem}\label{subsec:challenge}
The most common error-correction coding scheme is \emph{block coding},
in which an information sequence is divided into blocks of $k$ symbols each.
We denote a block using a $k$-tuple, $\bm{r}=(r_1,r_2,...,r_k)$.
For example, a 128-bit binary string can be divided into 16 blocks of 8-bit binary strings. In this case, $k=8$, and $r_i$ is either 0 or 1.
In general, $r_i$ can vary in other ranges of discrete symbols.
A fundamental requirement of error-correction coding
is that (i) the number of possible symbols for each $r_i$ must be the same,
and (ii) this number must be a prime power $p^m$, where $p$ is a prime number and $m$ is 
a positive integer. 
In abstract algebraic language, it requires the blocks to be in a \emph{Galois Field}, GF($p^m$). For example, the aforementioned 
8-bit strings are in GF(2).
Most of the current error-correction coding schemes (e.g., Reed-Solomon codes ~\cite{reed1960polynomial}) are built
on the algebraic operations in the polynomial ring of GF($p^m$).  At encoding
time, they transform $k$-tuple blocks into $n$-tuple blocks in the same Galois
field GF($p^m$), where $n$ is larger than $k$. At decoding time, some symbols
in the $n$-tuple blocks can be incorrect.  But as long as the total number of
incorrect symbols in a block is no more than
$\left\lfloor\frac{n-k}{2}\right\rfloor$ regardless of the locations of 
incorrections, the coding scheme can fully recover the
original $k$-tuple block.


In our problem, a text document consists of a sequence of letters. 
At first glance, we can divide the letter sequence into blocks of $k$ letters
each. Unfortunately, these blocks are ill-suited for traditional block coding schemes.
For example, consider a five-letter block, $(C_1,C_2,\mdots,C_5)$, with an original font $\bar{\bm{u}}$. Every letter
has a different capacity for embedding integers: in the codebook, 
$C_i$ has $s_i$ glyphs, so it can embed integers in the range $[0,s_i)$.
However, in order to use block coding, we need to find a prime power $p^m$ that
is no more than any $s_i,i=1...5$ (to construct a Galois field GF($p^m$)).
Then every letter can only embed $p^m$ integers, which can be significantly smaller than $s_i$. 
Perhaps a seemingly better approach is to find $t_i=\left\lfloor\log_2s_i\right\rfloor$ for every $s_i$,
and use the 5-letter block to represent a $T$-bit binary string, where $T=\sum_{i=1}^5 t_i$.
In this case, this binary string is in GF(2), valid for block coding. Still, this approach wastes 
much of the letters' embedding capacity.
For example, if a letter has 30 perturbed glyphs, this approach can only
embed integers in $[0,16)$.  As experimentally shown in \secref{sec:system},
it significantly reduces the amount of information that can be embedded.
In addition, traditional block coding method is often concerned with a noisy
communication channel, where an error occurs at individual \emph{bits}. In
contrast, our recognition error occurs at individual \emph{letters}.
When the glyph of a letter is mistakenly recognized, a chunk of bits becomes
incorrect, and the number of incorrect bits depends on specific letters.
Thus, it is harder to set a proper relative redundancy (i.e.,
$(n-k)/n$ in block coding) to guarantee successful error correction.


\subsection{Chinese Reminder Codes}\label{sec:crc}
To address these challenges, we introduce a new coding scheme based on a 1700-year old number theorem, 
the \emph{Chinese remainder theorem} (CRT) \cite{katz2007mathematics}.
Error-correction coding based on Chinese remainder theorem has been studied 
in the field of theoretical computing~\cite{goldreich1999chinese,boneh2000finding}, known as 
the \emph{Chinese Remainder Codes} (CRC). Building on CRC, we propose a coding scheme offering 
improved error-correction ability (\S\ref{sec:mld}). We start our presentation by 
defining our coding problem formally.

\paragraph{Problem definition}
Given a text document, we divide its letter sequence into blocks of $n$ letters each. 
Consider a block, denoted as $\mathbb{C}=(C_1,C_2,\mdots,C_n)$ in an original font $\bar{\bm{u}}$.
Our goal for now is to embed an integer $m$ in this $n$-letter block, 
where $m$ is in the range $[0,M)$.
We will map the plain message into a series of integers later in this section.
Formally, we seek an encoding function $\phi:m\to\bm{r}$, where $\bm{r}=(r_1,\mdots,r_n)$ is an $n$-vector of integers.
Following the coding theory terminology, we refer $\bm{r}$ as a \emph{codeword}.
The function $\phi$ needs to ensure that every $r_i$ can be embedded in the letter $C_i$.  Meanwhile, $\phi$ must be injective, 
so $\bm{r}$ can be uniquely mapped to $m$ through a decoding function $\phi^{+}:\bm{r}\to m$.  
Additionally, the computation of $\phi$
and $\phi^+$ must be fast, to encode and decode in a timely fashion.

\begin{figure}[t]
  \centering
  \includegraphics[width=0.72\columnwidth]{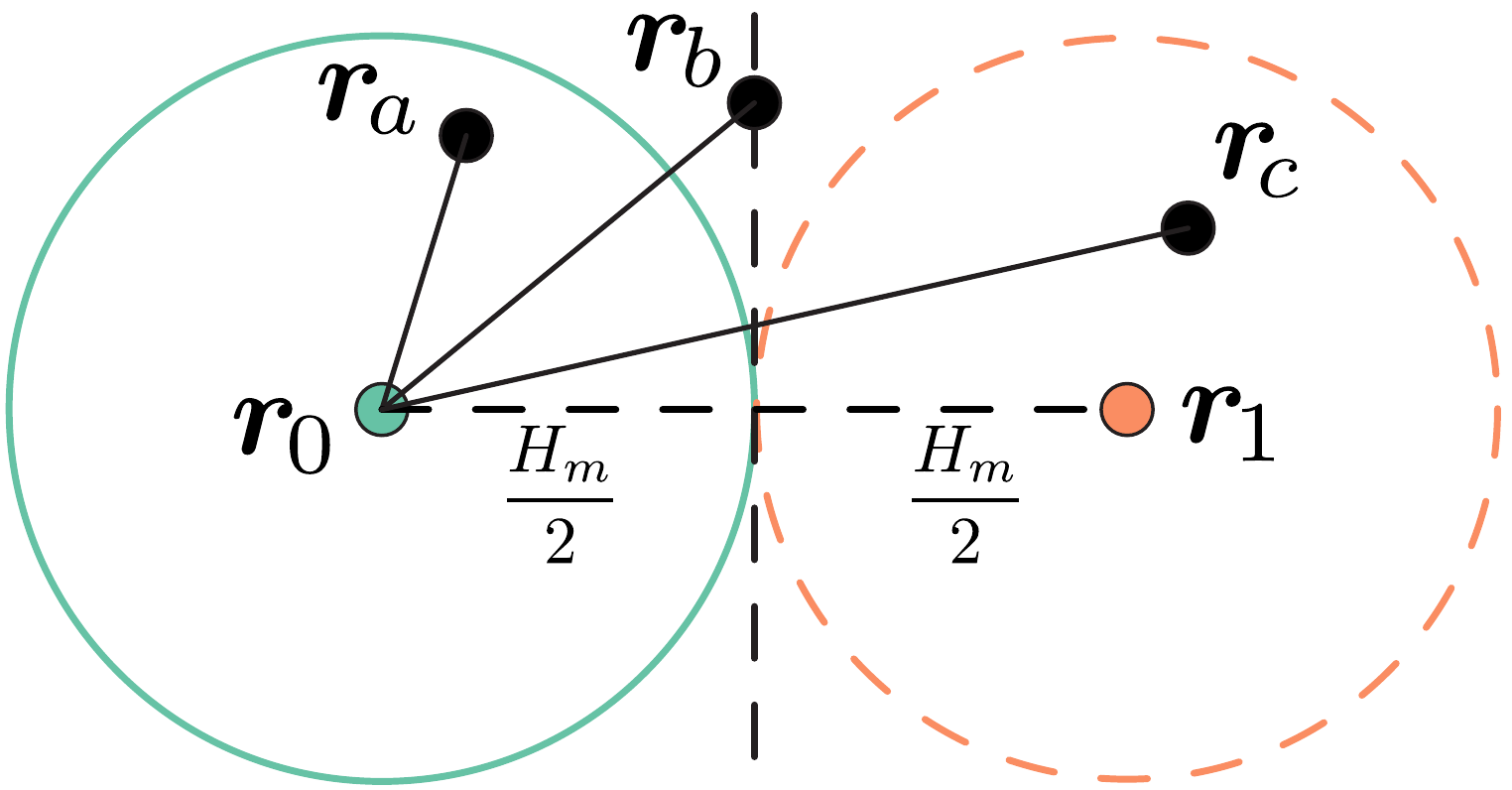}
  \vspace{-2mm}
  \caption{ {\bf Analogy} of Euclidean distance on 2D plane to gain intuition of Theorem~\ref{thm:B}.
  Here $\bm{r}_0$ and $\br_1$ represent two codewords having the minimum Hamming distance $H_m$.
  Suppose $\br_0$ is recognized as a code vector $\br$.
  If the distance $|\br-\br_0|<H_m/2$ (e.g., $\br_a$), 
  it is safe to decode $\br$ into $\br_0$, as no other codeword is closer to $\br$. 
  If $|\br-\br_0|>H_m/2$, there might be multiple codewords having
  the same distance to $\br$ (e.g., when $\br$ is $\br_b$), and $\br$ may be mistakenly decoded as $\br_1$ (e.g., when $\br$ is $\br_c$).
  } \label{fig:bound}
  \figspace
\end{figure}

\subsubsection{Hamming Distance Decoding}\label{subsec:hamming}
We now introduce \emph{Hamming Distance decoding}, a general decoding framework for block codes,
to pave the way for our coding scheme.
\vspace{1mm}
\begin{definition}[Hamming Distance]
Given two codewords $\bm{u}$ and $\bm{v}$,
the Hamming Distance $H(\bm{u},\bm{v})$ measures the number of pair-wise elements in which they differ. 
\end{definition}
\vspace{-1.8mm}
For example, given two codewords, $\bm{u}=(2,3,1,0,6)$ and $\bm{v}=(2,0,1,0,7)$,
$H(\bm{u},\bm{v})=2$.

Now we consider glyph recognition errors. 
If the integer retrieval from a letter's glyph is incorrect,
then the input $\tilde{\bm{r}}$ of the decoding function $\phi^+$ may not be a valid codeword.
In other words, because of the recognition errors, it is possible that no $m$ satisfies $\phi(m)=\tilde{\bm{r}}$.
To distinguish from a codeword, we refer to $\tilde{\bm{r}}$ as a \emph{code vector}.

The Hamming distance decoding uses a decoding function $\phi^+$ based on the
Hamming distance: $\phi^+(\tilde{\bm{r}})$ returns an integer $m$ such that the
Hamming distance $H(\phi(m),\tilde{\bm{r}})$ is minimized over all $m\in[0,M)$.
Intuitively, although $\tilde{\bm{r}}$ may not be a valid codeword, we 
decode it as the integer whose codeword is closest to $\tilde{\bm{r}}$ under
the measure of Hamming distance.
Let $H_m$ denote the \emph{minimum} Hamming distance among all pairs of valid codewords, namely,
\sLNM 
$$
H_m = \min_{\substack{i,j\in[0,M-1]\\i\ne j}}H(\phi(i),\phi(j)).
$$
\tLNM 
The error-correction ability of Hamming distance decoding is bounded by the
following theorem~\cite{lin2004error},
\vspace{1mm}
\begin{theorem}\label{thm:B}
    Let $E$ denote the number of incorrect symbols in a code vector $\tilde{\bm{r}}$. Then,
    the Hamming distance decoding can correctly recover the integer $m$ if $E\le\left\lfloor\frac{H_m-1}{2}\right\rfloor$; 
    it can detect errors in $\bar{\bm{r}}$ but not recover the integer correctly if $\left\lfloor\frac{H_m-1}{2}\right\rfloor<E<H_m$.
\end{theorem}
\vspace{-1mm}
An illustrative understanding of this theorem using an analogy of Euclidean
distance is shown in \figref{fig:bound}. This theorem holds regardless of the
encoding function $\phi$, as long as it is a valid injective function.  

\subsubsection{Chinese Reminder Codes}\label{subsec:CRC}
We now introduce the following version of the Chinese Reminder Theorem, whose proof can be found in~\cite{rosen2011elementary}.
\vspace{1mm}
\begin{theorem}[Chinese Remainder Theorem]
Let $p_1, p_2, ... p_k$ denote positive integers which are \textit{mutually prime} 
and $M=\prod_{i=1}^{k} p_i$.  
Then, there exists an injective function,
\sLNM 
$$
\phi: [0,M)\to [0,p_1)\times[0,p_2)\times...[0,p_k), 
$$ 
\tLNM 
defined as
$\phi(m) = (r_1,r_2,...,r_n)$, such that for all $m\in[0,M)$, $r_i=m\bmod p_i$.
\end{theorem}\vspace{-1mm}
This theorem indicates that given $k$ pairs of integers $(r_i,p_i)$ with all $p_i$ being mutually
prime, there exists a unique non-negative integer $m<\prod_{i=1}^{k} p_i$ satisfying $r_i=m\bmod p_i$
for all $i=1\mdots k$. Indeed, $m$ can be computed using the formula,
\begin{equation}\label{eq:computeA}
m=CRT(\bm{r},\bm{p})={r}_1b_1\frac{P}{p_1}+...+{r}_nb_n\frac{P}{p_n},
\end{equation}
where $P=\prod_{i=1}^{k}p_i$, and $b_i$ is computed by solving a system of modular equations,
\sLNM 
$$
b_i\frac{P}{p_i} \equiv 1\pmod{p_i}, \;\;i=1\mdots k,
$$
\tLNM 
using the classic Euclidean algorithm~\cite{rosen2011elementary}. 

If we extend the list of $p_i$ to $n$ mutually prime numbers ($n>k$) such that 
the $n-k$ additional numbers are all larger than $p_i$ up to $i=k$ (i.e., $p_j>p_i$ for any $j=k$+$1...n$ and $i=1...k$),
then we have an encoding function $\phi$ for non-negative integer $m<\prod_{i=1}^{k} p_i$:
\begin{equation}\label{eq:phi}\phi(m) =(m\bmod p_1, m\bmod p_2,...,m\bmod p_n).\end{equation}
This encoding function already adds redundancy:
because $m$ is smaller than the product of any $k$ numbers chosen from $p_i$, we 
can compute $m$ from any $k$ of the $n$ pairs of $(r_i,p_i)$, according to the Chinese Reminder Theorem.
Indeed, as proved in~\cite{goldreich1999chinese}, the minimum Hamming distance
of the encoding function~\eqref{eq:phi}
for all $0\le m<\prod_{i=1}^{k}p_i$ is $n-k+1$. Thus, the Hamming decoding 
function of $\phi$ can correct up to $\left\lfloor\frac{n-k}{2}\right\rfloor$ errors by Theorem~\ref{thm:B}. 

\paragraph{Encoding}
We now describe our encoding algorithm based on the encoding function $\phi(m)$ in \eqref{eq:phi}.

\begin{itemize}[itemsep=0pt,topsep=2pt,leftmargin=3.8mm]
\item\textbf{Computing $p_i$. }
Suppose that the letter sequence of a document has been divided into $N$ blocks, 
denoted as $\bb{C}_1,...,\bb{C}_{N}$, each with $n$ letters. Consider a block 
$\bb{C}_t=(C_1,...,C_n)$, where $C_i$ indicates a letter with its original font $\bm{u}_i$,
whose integer embedding capacity is $s_i$ (i.e., $C_i$'s font $\bm{u}_i$ has
$s_i$ perturbed glyphs in the codebook). We depth-first search $n$ mutually
prime numbers $p_i,i=1...n$, such that
$p_i\le s_i$ and the product of $k$ minimal $p_i$ is maximized. 
At the end, we obtain $p_i, i=1...n$ and the product of $k$ minimal $p_i$ denoted as $M_t$ for each block $\bb{C}_t$.
Note that if we could not find mutually prime numbers $p_i$ for block $\bb{C}_t$, we
simply ignore the letter whose embedding capacity is smallest among all $s_i$
and include $C_{n+1}$ to this block. We repeat this process until we find a valid
set of mutually prime numbers.
\item\textbf{Determining $m_t$. }
Given the plain message represented as a bit string $\bb{M}$, we now split the bits into a sequence of chunks,
each of which is converted into an integer and assigned to a block $\bb{C}_t$. 
We assign to each block $\bb{C}_t$ an integer $m_t$ with $\left\lfloor\log_2 M_t\right\rfloor$
bits, which is sequentially cut from the bit string $\bb{M}$ (see \figref{fig:cutoff}).

\begin{figure}[t]
  \centering
  \includegraphics[width=0.95\columnwidth]{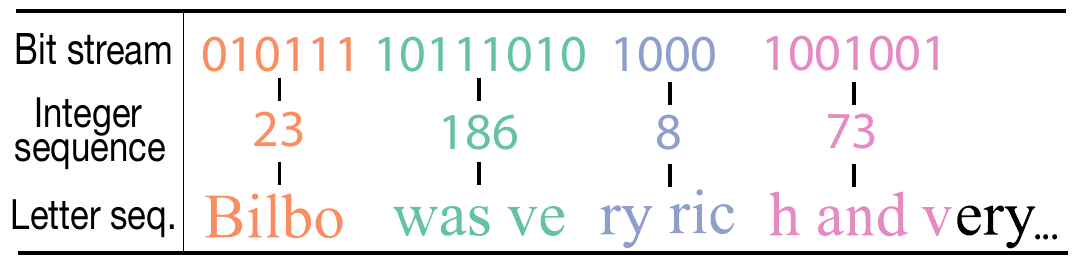}
  \vspace{-3mm}
  \caption{ {\bf An example} of determining $m_t$:
  (top) The bit string representation of the plain message;
  (bottom) Each block of letters is highlighted by a color;
  (middle) $m_t$ values assigned to each block.
  } \label{fig:cutoff}
  \figspace
\end{figure}

\item\textbf{Embedding.\; }For every block $\bb{C}_t$, we compute the codeword 
    using the CRT encoding function~\eqref{eq:phi}, obtaining $\bm{r}=(r_1,...,r_n)$. 
    Each $r_i$ is then embedded 
    in the glyph of the letter $C_i$ in the block $\bb{C}_t$ as described in \S\ref{subsec:embed}.
\end{itemize}

\paragraph{Decoding}
At decoding time, we recognize the glyphs of the letters in a document
and extract integers from them, as detailed in \secref{subsec:embed}. 
Next, we divide the letter sequence into blocks, 
and repeat the algorithm of computing $p_i$ and $M_t$ as in the encoding step, for every block. 
Given a block $\bb{C}_t$, the extracted integers from its letters form a code vector 
$\tilde{\bm{r}}_t=(\tilde{r}_1,...,\tilde{r}_n)$.
We refer to it as a code vector because some of the $\tilde{r}_i$ may be incorrectly recognized.
To decode $\tilde{\bm{r}}_t$, we first compute
$\tilde{m}_t=CRT(\tilde{\bm{r}}_t,\bm{p}_t)$ where $\bm{p}_t$ stacks all $p_i$ in the block. If
$\tilde{m}_t<M_t$, then 
$\tilde{m}_t$ is the decoding result $\phi^+(\tilde{\bm{r}}_t)$, 
because the Hamming distance $H(\phi(\tilde{m}_t),\tilde{\bm{r}})=0$.
Otherwise,
we decode $\tilde{\bm{r}}_t$ using the Hamming decoding function: 
concretely, since we know the current block can encode an integer in the range $[0,M_t)$,
we decode $\tilde{\bm{r}}_t$ into the integer $m_t$ by finding
\begin{equation}\label{eq:decode}
    m_t = \phi^+(\tilde{\bm{r}}) = \arg\min_{m\in[0,M_t)} H(\phi(m),\tilde{\bm{r}}).
\end{equation}
As discussed above, this decoding function can correct up to
$\left\lfloor\frac{n-k}{2}\right\rfloor$ incorrectly recognized
glyphs in each block.  Lastly, we convert $m_t$ into a bit string and
concatenate $m_t$ from all blocks sequentially to recover the plain message.

\paragraph{Implementation details}
A few implementation details are worth noting. First, oftentimes letters in a document can carry a bit string 
much longer than the given plain message. To indicate the end of the message, 
we attach a special chunk of bits (end-of-message bits) at the end of each plain message, very much akin 
to the end-of-line (newline) character used in digital text systems.
Second, in practice, blocks should be relatively short (i.e., $n$ is small). If $n$ is large,
it becomes much harder to find $n$ mutually prime numbers that are no more than each letter's embedding capacity.
In practice, we choose $n=5$ and $k=3$ which allows one mistaken letter in every 5-letter block.
The small $n$ and $k$ also enable brute-force search of $m_t$ in~\eqref{eq:decode} sufficiently fast,
although there also exists a method solving~\eqref{eq:decode} in polynomial time~\cite{goldreich1999chinese,boneh2000finding}.

\subsection{Improved Error Correction Capability}\label{sec:mld}
Using the Chinese Remainder Codes, the error-correction capacity is
upper bounded. 
Theorem~\ref{thm:B} indicates that
at most $\left\lfloor\frac{n-k}{2}\right\rfloor$ mistakenly recognized
glyphs are allowed in every letter block.
We now break this theoretical upper bound to further improve our error-correction ability,
by exploiting specific properties of our coding problem.
To this end, we propose a new algorithm 
based on the maximum likelihood decoding~\cite{lin2004error}.

In coding theory, maximum likelihood decoding is not an algorithm. Rather, it is
a decoding philosophy, a framework that models the decoding process from a probabilistic 
rather than an algebraic point of view.
Consider a letter block $\bb{C}$ and the code vector $\tilde{\bm{r}}$ formed by the extracted integers from $\bb{C}$.
We treat the true codeword $\bm{r}$ encoded by $\bb{C}$ as a \emph{latent variable} (in statistic language),
and model the probability of $\bm{r}$ given the extracted code vector $\tilde{\bm{r}}$, namely $\bb{P}(\bm{r}|\tilde{\bm{r}})$.
With this likelihood model,
our decoding process finds a codeword $\bm{r}$ that maximizes the probability $\bb{P}(\bm{r}|\tilde{\bm{r}})$,
and decodes $\bm{r}$ into an integer using the Chinese Remainder Theorem formula~\eqref{eq:computeA}.

When there are at most $\left\lfloor\frac{n-k}{2}\right\rfloor$ errors in a block, the Hamming
decoding function~\eqref{eq:decode} is able to decode. In fact,
it can be proved that when the number of errors is under this bound, 
there exists a unique $m\in[0,M_t)$ that minimizes $H(\phi(m),\tilde{\bm{r}})$~\cite{lin2004error},
and that when the number of errors becomes
$\left\lfloor\frac{n-k}{2}\right\rfloor+1$, 
there may be multiple $m\in[0,M_t]$
reaching the minimum (see \figref{fig:bound} for an intuitive explanation). 
The key idea of breaking this bound is to use a likelihood model to choose an $m$ when 
ambiguity occurs for the Hamming decoding function~\eqref{eq:decode}.

\begingroup
\setlength{\columnsep}{0pt} %
 \begin{wrapfigure}[9]{r}{0.52\columnwidth}
 \vspace{-3mm}
 \centering
 \includegraphics[width=\linewidth]{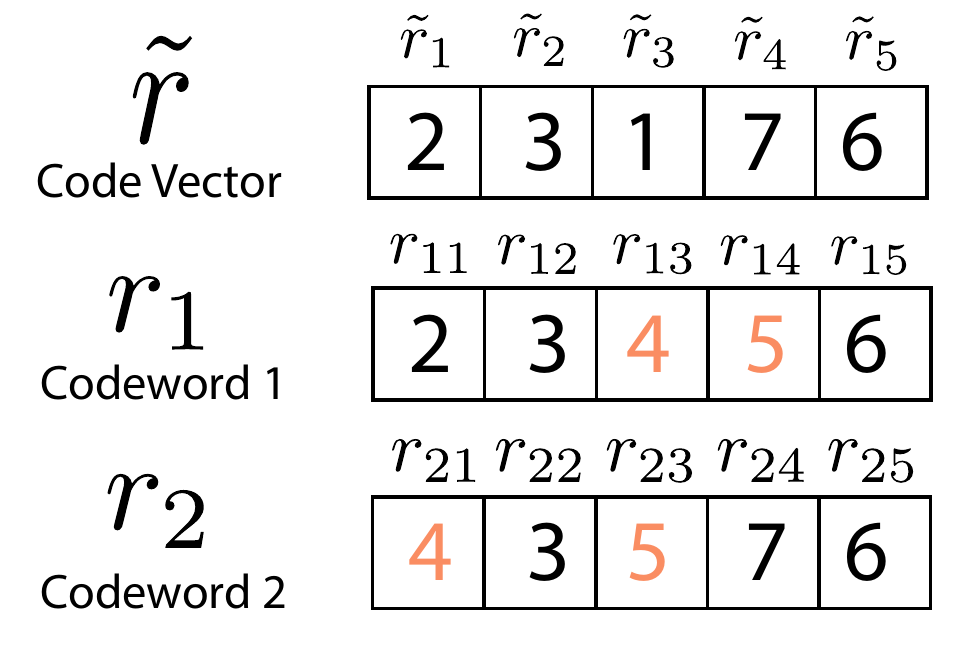}
 \vspace{-2mm}
 \end{wrapfigure}
Consider a block with $\left\lfloor\frac{n-k}{2}\right\rfloor+1$ errors,
and suppose in~\eqref{eq:decode} we find $N_c$ different integers $m_i$, all of which lead to the same minimal 
Hamming distance to the code vector $\tilde{\bm{r}}$. Let $\bm{r}_i=\phi(m_i),i=1\mdots N_c$, denote the corresponding codewords.
We use another subscript $j$ to index the integer elements in code vectors and codewords.
For every $\bm{r}_i$, some of its integer elements $r_{ij}$ differs from the corresponding
integer elements $\tilde{r}_j$. 
Here $r_{ij}$ can be interpreted as the index of the perturbed glyphs of the $j$-th letter.
We denote this glyph as $\bm{u}_{ij}$ and the letter's glyph 
extracted from the input image as $\bm{f}$.
If $r_{ij}$ is indeed the embedded integer,
then $\bm{f}$ is mistakenly recognized, resulting in a different glyph number $\tilde{r}_j$. 

We first model the probability of this occurrence, denoted as $\bb{P}(\tilde{r}_j|r_{ij})$, 
using our ``distance'' metric.
Intuitively, the closer the two glyphs are, the more
likely one is mistakenly recognized as the other. Then the probability of
recognizing a codeword $\bm{r}_i$ as a code vector
$\tilde{\bm{r}}$ accounts for all inconsistent element pairs in $\bm{r}_i$ and
$\tilde{\bm{r}}$, namely,
\begin{equation}\label{eq:prob}
\bb{P}(\tilde{\bm{r}}|\bm{r}_i) = \prod_{j,r_{ij}\ne\tilde{r}_j}
\frac{g(\bm{u}_{ij},\bm{f})}{\sum_{k=1}^{s_{j}}
g(\bar{\bm{u}}_{j}^k,\bm{f})}.
\end{equation}
Here $g(\cdot,\cdot)$ is (inversely) related to our distance metric between two glyphs:
for pixel images, $g(\bm{u}_{ij},\bm{f})$ is the probability of recognizing $\bm{f}$ as $\bm{u}_{ij}$, 
which is the $r_{ij}$-th softmax output from the CNN of the $i$-th letter, 
given the input glyph image $\bm{f}$ (recall \S\ref{subsec:architecture});
for vector graphics, $g(\bm{u}_{ij},\bm{f})=1/d(\bm{u}_{ij},\bm{f})$, the inverse of the 
vector graphic glyph distance (defined in \S\ref{subsec:embed}).
The denominator is for normalization, where $\bar{\bm{u}}_{j}^k$ iterates through all
perturbed glyphs of the $j$-th letter in the codebook. For pixel images, the denominator 
is always 1 because of the unitary property of the softmax function.
\endgroup
Lastly, the likelihood $\bb{P}(\bm{r}_i|\tilde{\bm{r}})$ needed for decoding 
is computed by Bayes Theorem,
\begin{equation}\label{eq:prob2}
\bb{P}(\bm{r}_i|\tilde{\bm{r}}) = \frac{\bb{P}(\tilde{\bm{r}}|\bm{r}_i)\bb{P}(\bm{r}_i)}{\bb{P}(\tilde{\bm{r}})}.
\end{equation}
Here, $\bb{P}(\tilde{\bm{r}})$ is fixed, and so is $\bb{P}(\bm{r}_i)$ as all codewords are equally likely to be used.
As a result, we find $\bm{r}_i$ that maximizes \eqref{eq:prob2} among all $N_c$ ambiguous codewords, and decode $\tilde{\bm{r}}$
using $\bm{r}_i$.

In our experiments (\secref{sec:valid}), we show that the proposed decoding scheme indeed improves the error-correction
ability. For our implementation wherein $n=5$ and $k=3$, in each block we are able to correct
up to \textbf{2} errors, as opposed to 1 indicated by Theorem~\ref{thm:B}, 
thereby increasing the error tolerance from 20\% to 40\%.

\begin{figure}[t]
  \centering
  \includegraphics[width=0.99\columnwidth]{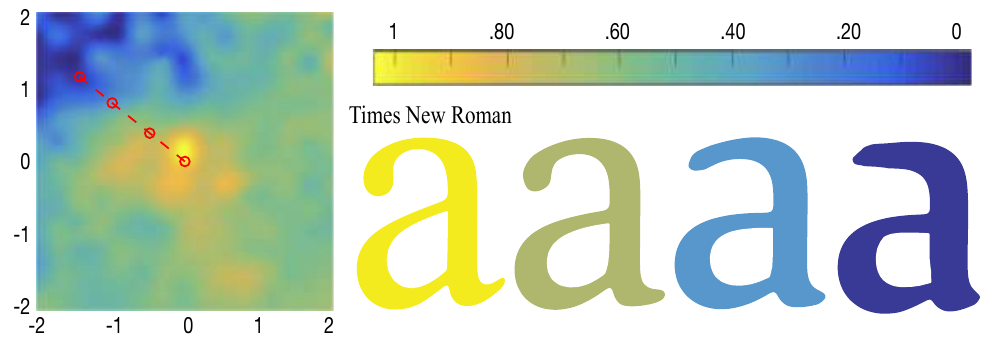}
  \vspace{-3.5mm}
  \caption{ {\bf MTurk result.}  
  (left) The font manifold of the letter ``a'' color-mapped using the perceptual similarity value of each point
  to the standard Time New Roman font.
  (right) Sampled glyphs that correspond to the manifold locations indicated by red circles.
  The colors indicate their perceptual similarity to the Times New Roman.
  } \label{fig:mturkRet}
  \figspace
\end{figure}
\section{Perceptual Evalutation}\label{sec:perceptual}
We now describe our crowd-sourced perceptual studies on Mechanical Turk (MTurk).
The goal of the studies is, for each character in a particular font, to 
find a set of glyphs that are perceptually similar to its original glyph $\bar{\bm{u}}$. 
When constructing the glyph codebook, 
we use these sets of glyphs to initialize the candidates of perturbed glyphs (recall \S\ref{subsec:codebook}).

The user studies adopt a well-established protocol: we present the MTurk raters
multiple questions, and each question uses a two-alternative forced choice
(2AFC) scheme, i.e., the MTurk raters must choose one from two options.  We
model the rater's response using a logistic function (known as the Bradley-Terry
model~\shortcite{bradley1952rank}), which allows us to learn a perceptual
metric (in this case the perceptual similarity of glyphs) from the raters'
response. 
We note that this protocol has been used previously in other perceptual studies (e.g.,~\cite{ODonovan2014,Um2017Perceptual}),
and that we will later refer back to this protocol in \S\ref{sec:valid} when we evaluate the perceptual quality of our results. 
Next, we describe the details of the studies. 



\begin{figure}[h]
  \centering
  \vspace{-2mm}
  \includegraphics[width=0.95\columnwidth]{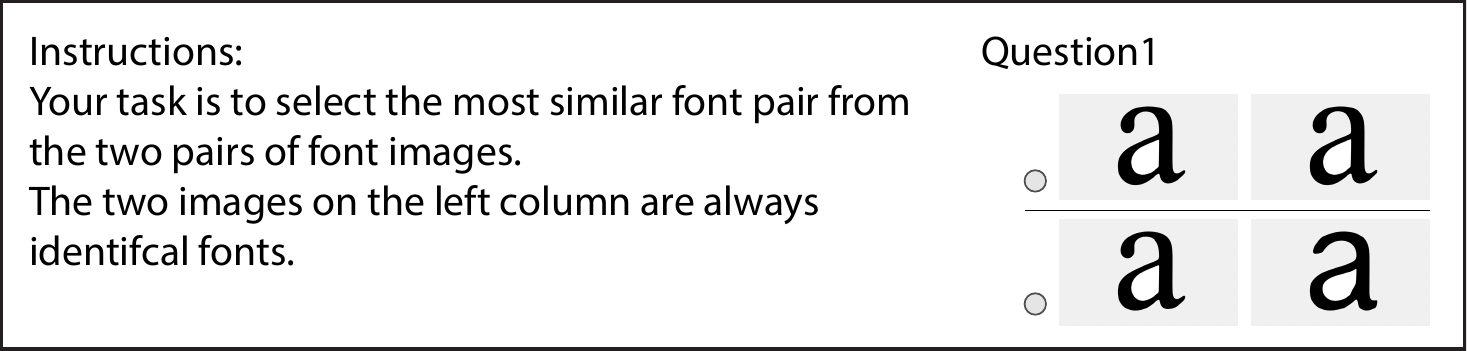}
  \vspace{-3mm}
  \caption{ MTurk user interface. } \label{fig:mturk}
  \vspace{-2mm}
  \figspace
\end{figure}

\paragraph{Setup}
We first choose on the font manifold a large region centered
at $\bar{\bm{u}}$, and sample locations densely in this region.
In practice, all font manifolds are in 2D, and we choose 400
locations uniformly in a squared region centered at $\bar{\bm{u}}$. Let $\mathcal{F}$ denote
the set of glyphs corresponding to the sampled manifold locations. Then, in each MTurk question,
we present the rater a pair of glyphs randomly selected from $\mathcal{F}$, and ask
which one of the two glyphs looks closer to the glyph of the original font $\bar{\bm{u}}$.
An example is shown in \figref{fig:mturk}.
We assign 20 questions to each MTurk rater. Four of them are control questions,
which are designed to detect untrustworthy raters. The four questions 
ask the rater to compare the same pair of glyphs presented in different order.
We reject raters whose answers have more than one inconsistencies among the control questions.
At the end, about 200 raters participated the studies for each character.

\paragraph{Analysis}
From the raters' response, we estimate a scalar value
$s_i=s(\bm{u}_i;\bar{\bm{u}})$, the perceptual similarity value of the glyph
$\bm{u}_i$ to the original glyph $\bar{\bm{u}}$, for every glyph
$\bm{u}_i\in\mathcal{F}$.
Let a set of tuples $\mathcal{D}=\{(\bm{u}_i,\bm{u}_j,u,q)\}$
record the user study responses,
where $\bm{u}_i$ and $\bm{u}_j$ are randomly selected glyphs
being compared, $u$ is the MTurk rater's ID, and the binary value $q$ is the rater's choice:
$q=1$ indicates the rater judges $\bm{u}_i$ perceptually closer to $\bar{\bm{u}}$,
and $q=0$ otherwise.
We model the likelihood of the rater's choice using a logistic function,
\begin{equation}\label{eq:tb}
p(q=1|\bm{u}_i,\bm{u}_j,u) = \frac{1}{1 + \exp\left( r_u (s_i - s_j)\right)},
\end{equation}
where the scalar $r_u$ indicates the rater's reliability.
With this model, all the similarity values $s_i$ and user reliability values $r_u$ are obtained by 
minimizing a negative log-likelihood objective function,
\begin{equation*}
\begin{split}
E(\bm{s},\bm{r}) =& -\sum_{k}\left[ q^k\ln p(q=1|\bm{u}^k_i,\bm{u}^k_j,u^k)\; + \right.\\
& \relphantom{\sum_k}\left.(1-q^k)\ln p(q=0|\bm{u}^k_i,\bm{u}^k_j,u^k)\right],
\end{split}
\end{equation*}
where $k$ indices the MTurk response in $\mathcal{D}$, 
$\bm{r}$ stacks the raters' reliability values, and
$\bm{s}$ stacks the similarity values for all $\bm{u}_i\in\mathcal{F}$.
The $s_i$ values are then normalized in the range of $[0,1]$.

We learn the similarity values for the glyphs of 
every character with a font $\bar{\bm{u}}$ independently.
\figref{fig:mturkRet} visualizes the perceptual similarity 
of glyphs near the standard Times New Roman for the character ``a''. 
Lastly, We iterate through all the pre-sampled glyphs $\bm{u}_i\in\mathcal{F}$,
and add those whose similarity value $s_i$ is larger than a threshold (0.85 in
practice) into a set $\mathcal{C}$, forming the set of perceptually similar
glyphs for constructing the codebook (in \S\ref{subsec:codebook}).

\section{Applications}\label{sec:app}
Our method finds many applications.
In this section, we discuss {four} of them,
while referring to the supplemental video for their demonstrations 
and to \secref{subsec:ret} for implementation summaries.

\subsection{Application I: Format-Independent Metadata}\label{subsec:meta}
Many digital productions carry \emph{metadata} that provide additional 
information, resources, and digital identification~\cite{greenberg2005understanding}.
Perhaps most well-known is the metadata embedded in photographs, providing
information such as camera parameters and copyright. 
PDF files can also contain metadata\footnote{If you are 
reading this paper with Adobe Acrobat Reader, you can view its metadata by choosing ``File$\to$Properties''
and clicking the ``Additional Metadata'' under the ``Description'' tab.}.
In fact, metadata has been widely used by numerous tools to edit and organize digital files. 

Currently, the storage of metadata is \emph{ad hoc}, depending on specific file format.
For instance, a JPEG image stores metadata in its EXIF header,
while an Adobe PDF stores its metadata in XML format with Adobe's XMP framework~\cite{adobemanager}.
Consequently,
metadata is lost whenever one converts an image from JPEG to PNG format, or rasterizes
a vector graphics image into a pixel image. Although it is possible to develop a careful
converter that painstakingly preserves the metadata across all file formats,
the metadata is still lost whenever the image or document is {printed} on paper.

Our FontCode technique can serve as a means to host text document metadata.
More remarkably, the metadata storage in our technique is \emph{format-independent},
because format conversion and rasterization of text documents, as well as printing on paper,
all preserve the glyphs of letters.
Once information is embedded in a document, one can freely convert it to a different file format, or rasterize
it into a pixel image (as long as the image is not severely downsampled),
or print it on a piece of paper. Throughout, the metadata is retained (see video).


\subsection{Application II: Imperceptible Optical Codes}
Our FontCode technique can also be used as optical barcodes embedded in a text document,
akin to QR codes~\cite{denso2011qr}. Barcodes have numerous applications in advertising, 
sales, inventory tracking, robotics, augmented reality, and so forth.
Similar to QR codes that embed certain level of redundancy to correct decoding error,
FontCode also supports error-correction decoding.
However, all existing barcodes require to print black-and-white blocks and bars, which can be
visually distracting and aesthetically imperfect. 
Our technique, in contrast, enables not only an optical code but an unobtrusive optical code,
as it only introduces subtle changes to the text appearance.
Our retrieval algorithm is sufficiently fast to provide point-and-shoot
kind of message decoding.
It can be particularly suitable for use as a replacement of QR codes in an artistic work such
as a poster or flyer design, where visual distraction needs to be minimized.
As a demonstration, we have implemented an iPhone application to read coded text (see video).


\subsection{Application III: Encrypted Message Embedding}\label{subsec:enc}
Our technique can further encrypt a message when embedding it in a document, even if the entire
embedding and retrieval algorithms are made public.
Recall that when embedding an integer $i$ in a letter $c$ of a glyph $\bar{\bm{u}}$,
we replace $\bar{\bm{u}}$ with a glyph chosen from a list of perturbed glyphs in
the codebook. Let $\mathbb{L}_c=\{\bm{u}_0,\ldots,\bm{u}_{N_c-1}\}$ denote this list.
Even though the perturbed glyphs for every character in a particular font are precomputed, 
the order of the glyphs in each list $\mathbb{L}_c$
can be arbitrarily user-specified. 
The particular orders of all $\mathbb{L}_c$ together can serve as an encryption key.

For example, when Alice and Bob\footnote{Here we 
follow the convention in cryptography, using Alice and Bob as placeholder names
for the convenience of presenting algorithms.}
communicate through encoded documents,
they can use a publicly available codebook, but agree on a private key, which specifies
the glyph permutation of each list $\mathbb{L}_c$. If an original list $\{\bm{u}_0,\bm{u}_1,\bm{u}_2,\ldots\}$
of a letter is permuted into $\{\bm{u}_{p_0},\bm{u}_{p_1},\bm{u}_{p_2},\ldots\}$ by the key,
then Alice uses the glyph $\bm{u}_{p_i}$, rather than $\bm{u}_i$, to embed an integer $i$ in the letter,
and Bob deciphers the message using the same permuted codebook.
For a codebook that we precomputed for Times New Roman (see the supplemental document), 
if we only consider lowercase English alphabet, 
there exist $1.39\times10^{252}$ different keys (which amount to a 837-bit key);
if we also include uppercase English alphabet,
there exist $5.73\times10^{442}$ keys (which amount to a 1470-bit key).
Thus, without resorting to any existing cryptographic algorithm, our method already offers a 
basic encryption scheme. Even if others can 
carefully examine the text glyphs and discover 
that a document is indeed embedding a message, the message can still be protected from leaking.

\subsection{Application IV: Text Document Signature}
Leveraging existing cryptographic techniques, 
we augment FontCode to propose a new digital signature technique, 
one that can authenticate the source of a text document and 
guarantee its integrity (thereby protecting from tampering).
This technique has two variations, working as follows:

\paragraph{Scheme 1}
When Alice creates a digital document, she maps the document content (e.g., including
letters, digits, and punctuation)
into a bit string through a cryptographic hash function such as
the MD5~\cite{Rivest:1992:MMA} and SHA~\cite{Eastlake2001USH}.
We call this bit string the document's \emph{hash string}.
Alice then chooses a private key to permute the codebook as described in \S\ref{subsec:enc},
and uses the permuted codebook to embed the hash string into her document.
When Bob tries to tamper this document, any change leads to 
a different hash string. Without knowing Alice's private key, he cannot embed the 
new hash string in the document, and thus cannot tamper the document successfully.
Later, Alice can check the integrity of her document, by extracting the embedded hash string
and comparing it against the hash string of the current document.

\paragraph{Scheme 2}
The above algorithm allows only Alice to check the integrity of her document, as only 
she knows her private key. By combining with \emph{asymmetric cryptography} 
such as the RSA algorithm~\cite{rivest1978method}, we allow everyone 
to check the integrity of a document but not to tamper it.
Now, the codebook is public but not permuted. 
After Alice generates the hash string of her document, she encrypts the hash string
using her private key by an asymmetric cryptosystem, and obtains an \emph{encrypted string}. 
Then, she embeds the encrypted string in the document using the FontCode method,
while making her public key available to everyone.
In this case, Bob cannot tamper the document, as he does not have Alice's private key 
to encrypt the hash string of an altered document.
But everyone in the world can extract the encrypted string using the FontCode method,
and decipher the hash string using Alice's public key.  
If the deciphered hash string matches the hash string of the current document, it proves
that (i) the document is indeed sourced from Alice, and (ii) the document has not been modified by others.

\paragraph{Advantages}
In comparison to existing digital signatures such as those in Adobe PDFs,
our method is \emph{format-independent}. 
In contrast to PDF files whose signatures are lost when the files are rasterized or printed on physical papers,
our FontCode signature is preserved regardless of file format conversion, rasterization, and physical printing.

\paragraph{Further extension}
In digital text forensics, it is often desired to not only detect tampering but also
locate where in the text the tampering occurs.
In this regard, our method can be extended for more detailed tampering detection.
As shown in our analysis (\S\ref{sec:system}), in a typical English document,
we only need about {80} letters to embed (with error correction) 
a string of 128 bits, which is the length of a hash string resulted 
from a typical cryptographic hash function (e.g., MD5).
Given a document, we divide its text into a set of segments, each with at least {80} letters.
We then compute a hash string for each segment and embed the encrypted strings in individual segments.
This creates a fine granularity of text signatures, allowing the user to check
which text segment is modified, and thereby locating tampering occurrences more
precisely. For example, in the current text format of this paper, every
two-column line consists of around 100 letters, meaning that our method can
identify tampering locations up to two-column lines in this paper.
To our knowledge, digital text protection with such a fine granularity has not been realized.

\paragraph{Discussion about the storage}
We close this section with a remark on the memory footprint of our steganographic
documents. If a document is stored as a pixel image, then it consumes no
additional memory.  If it is in vector graphics format, our current
implementation stores the glyph contour polylines of all letters.
A document with 376 letters with 639 bits encoded will consume 1.2M memory in a
compressed SVG form and 371K in a compressed JPEG format. 
PDF files can embed glyph shapes in the file and refer to those glyphs in the text.
Using this feature, we can also embed the entire codebook in a PDF, introducing about 1.3M storage overhead, 
regardless of the text length.
In the future, if all glyphs in the codebook are pre-installed on the operating
system, like the current standardized fonts, the memory footprint of vector
graphic documents can be further reduced.

\begin{figure}[t]
 \centering
 \includegraphics[width=0.99\columnwidth]{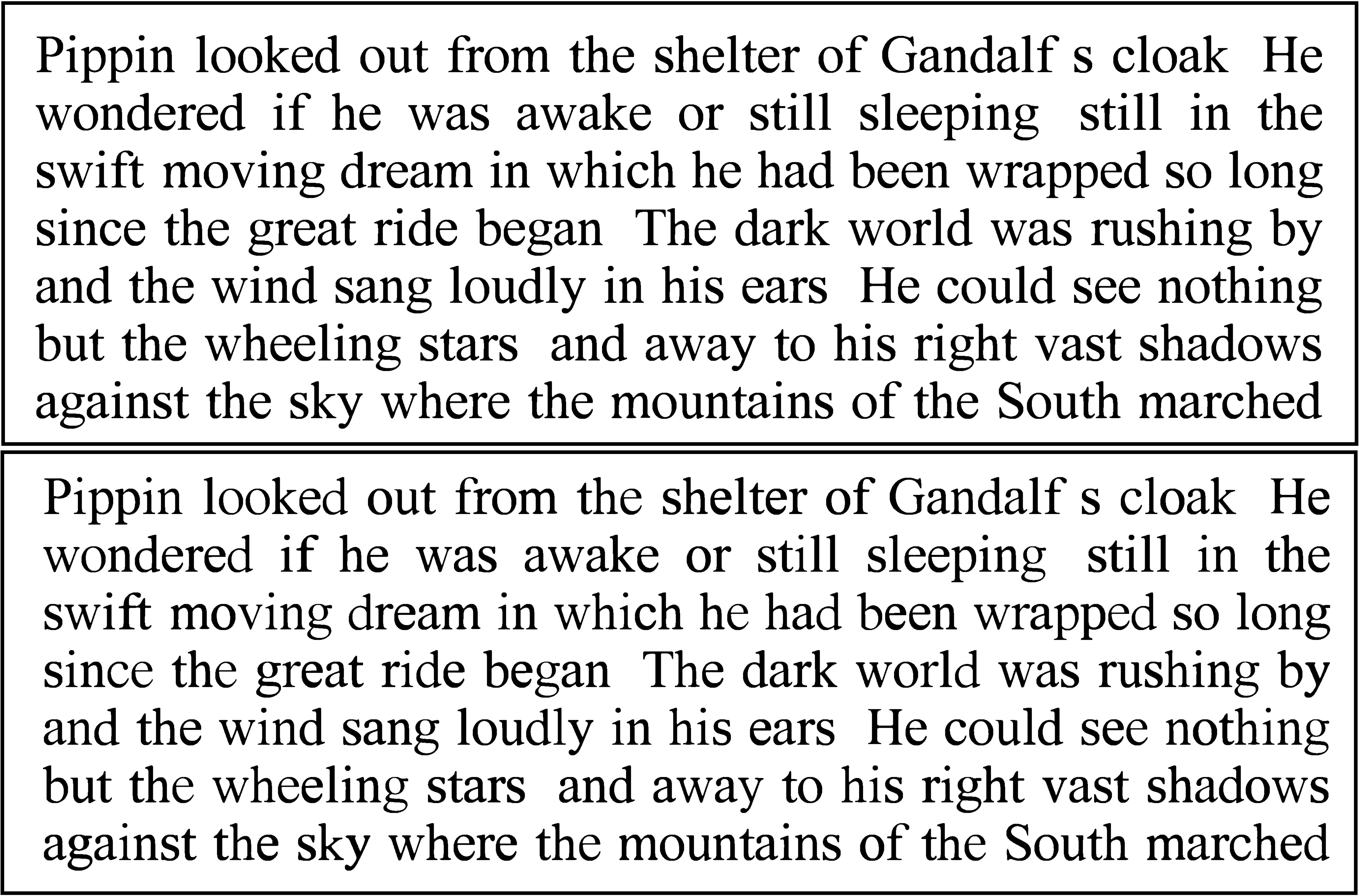}
 \vspace{-3mm}
 \caption{
 {\bf (top)} an input text document.
 {\bf (bottom)} the output document that embeds a randomly generated message.
 \label{fig:sample_result}}
 \vspace{-4mm}
\end{figure}
\section{Results and Validation}\label{sec:results}
We now present the results and experiments to analyze the performance of
our technique and validate our algorithmic choices.
Here we consider text documents with English alphabet, including 
both lower- and upper-case letters, 
while the exact method can be directly applied to digits and other special characters. 
We first present our main results (\secref{subsec:ret}), 
followed by the numerical (\secref{sec:system}) and perceptual (\secref{sec:valid}) evaluation of our method.

\subsection{Main Results}\label{subsec:ret}
We implemented the core coding scheme on a Intel Xeon E5-1620 8 core 3.60GHz
CPU with 32GB of memory. The CNNs are trained with an NVidia Geforce TITAN X GPU. 
Please see our accompanying video for the main results.  A side-by-side
comparison of an original document with a coded document is shown in
\figref{fig:sample_result}.

\paragraph{Metadata viewer} 
We implemented a simple text document viewer that loads 
a coded document in vector graphics or pixel image format.
The viewer displays the document. Meanwhile, it extracts the embedded metadata with
our decoding algorithm and presents it in a side panel.

\paragraph{Unobtrusive optical codes} 
We also implemented an iPhone application (see \figref{fig:teaser}), by which
the user can take a photo of an encoded text displayed 
on a computer screen or printed on paper.  The iPhone
application interface allows the user to select a text region to capture.  The
captured image is sent through the network to a decoding server, which
recovers the embedded message and sends it back to the smartphone.

\paragraph{Embedding encrypted message}
Our implementation allows the user to load an encryption key file that
specifies the permutation for all the lists of perturbed glyphs in the
codebook.  The permutation can be manually edited, or randomly
generated---given a glyph list of length $n$, one can randomly sample a
permutation from the permutation group
$\mathbb{S}_n$~\cite{seress2003permutation} and attach it to the key.

\paragraph{Text document signature}
We use our technique to generate a MD5 signature as described in \secref{sec:app}.
Since the MD5 checksum has only 128 bits, we always embed it in letters from 
the beginning of the text. Our text document viewer can check the signature and 
alert the user if the document shows as tampered.


\subsection{Validation}\label{sec:system}


\begin{figure}[t]
 \centering
 \includegraphics[width=0.92\columnwidth]{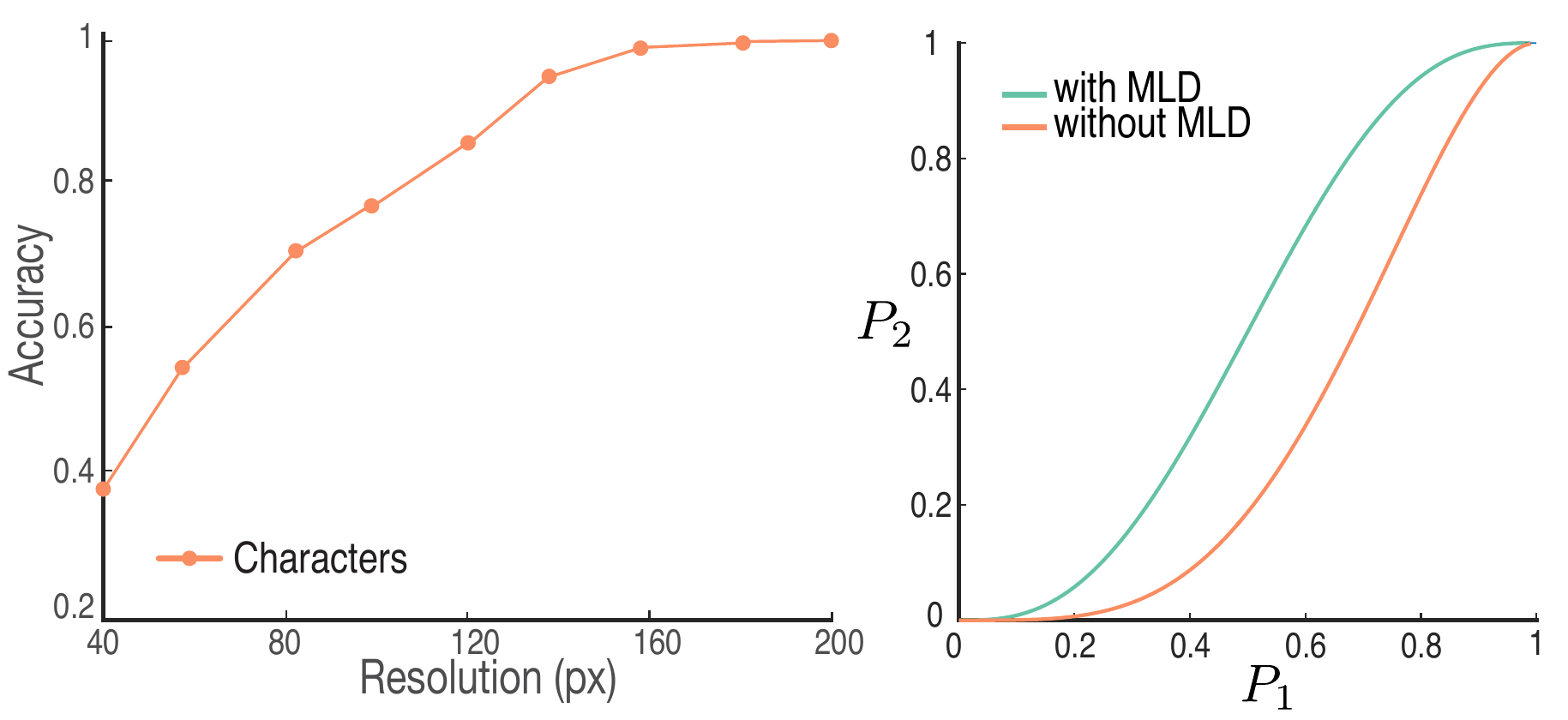}
 \vspace{-4.5mm}
 \caption{{\bf (left)} The accuracy of our CNN decoder 
 changes as the resolution in height of the letters increases.
 {\bf (right)} Theoretical improvement of maximum likelihood decoding (green)
 over the Chinese Remainder coding (orange).
 \label{fig:maximprove}}
 \vspace{-4mm}
\end{figure}

\paragraph{Information capacity}
As described in \secref{sec:coding}, we use $n=5$ and $k=3$ in our 
error-correction coding scheme. In every 5-letter block,
if the mutually prime numbers are $p_i,i=1\mdots5$, then this block can
encode integers in $[0, p_1p_2p_3)$, where $p_1$, $p_2$, and $p_3$ are
the smallest three numbers among all $p_i$. Thus, this block can encode
at most $\left\lfloor\log_2(p_1p_2p_3)\right\rfloor$ bits of information.


To estimate the information capacity of our scheme for English text, we
randomly sample 5 characters from the alphabet to forming a block.  The
characters are sampled based on the widely 
known English letter frequencies (e.g., ``e'' is the most frequently used while ``z'' is 
the least used)~\cite{ferguson2003practical}. We compute the average number of bits that 
can be encoded by a character. The result is \textbf{1.77},
suggesting that on average we need 73 letters to encode a 128-bit MD5 checksum
for the application of digital signature (\secref{sec:app}).

\begingroup
\setlength{\columnsep}{4pt}
\begin{wrapfigure}[14]{r}{0.33\columnwidth}
 \centering
 \vspace{-4mm}
 \includegraphics[width=\linewidth]{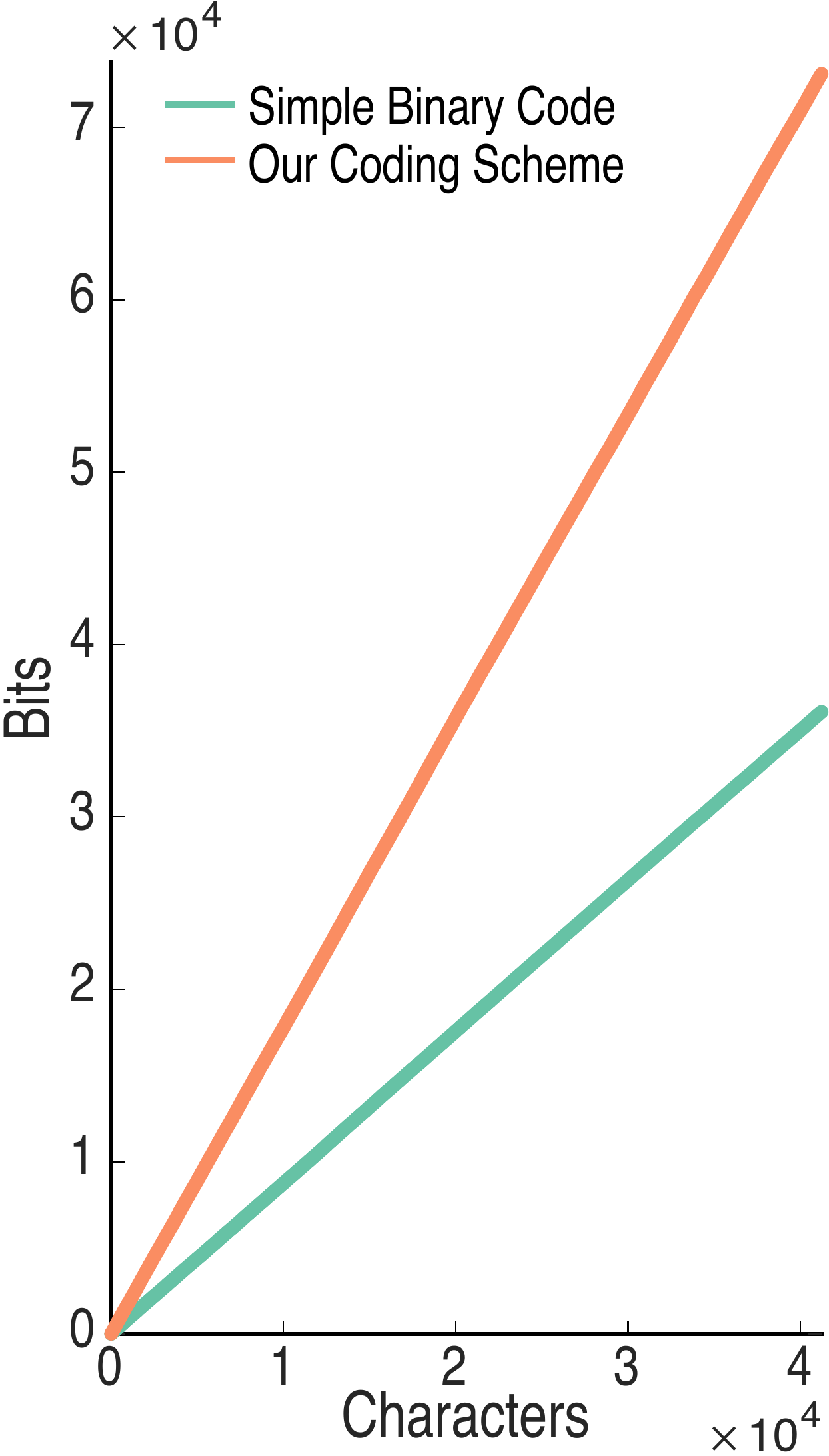}
\end{wrapfigure}
Next, we compare our coding scheme with the simple approach discussed in \secref{subsec:challenge}.
Recall that our method can correct at least one error in a block of 5 letters, which
amounts to correcting $\log_2(\max s_i)$ bits out of the total $\sum_{i=1}^5\log_2 s_i$ bits.
The simple approach in \secref{subsec:challenge} can store $\sum_{i=1}^5\left\lfloor\log_2 s_i\right\rfloor$
bits of information. But in order to correct one recognition error of the letter with 
the standard linear block codes, it needs to spend $2\left\lceil\log_2(\max s_i)\right\rceil$
bits for adding redundancy, leaving $\sum_{i=1}^5\left\lfloor\log_2 s_i\right\rfloor-2\left\lceil\log_2(\max s_i)\right\rceil$
bits to store information.
We compare it with our method using \emph{The Lord of The Rings, Chapter 1} as our input text, which contains in total
41682 useful letters (9851 words). 
As shown in the adjacent figure, as the number of letters increases,
our method can embed significantly more information.

\paragraph{Decoding accuracy}
We first evaluate the glyph recognition accuracy of our CNNs. For every
character, we print it repeatedly on a paper with randomly chosen glyphs from the codebook
and take five photos under different lighting conditions. Each photo has
regions of around 220px$\times$220px containing a character. We use these photos to test CNN recognition 
accuracy, and for all characters, the accuracy is above 90\%.

We also evaluate the decoding accuracy of our method.  
We observed that decoding errors are mainly caused by image rasterization.
If the input document is in vector graphics format, the decoding result is fully accurate, as we know the glyph outlines precisely. 
Thus, we evaluate the decoding accuracy with pixel images.
We again use \emph{The Lord of The Rings, Chapter 1} to encode a random bit string. We rasterize
the resulting document into images with different resolutions, and 
measure how many letters and blocks can be decoded correctly.
\figref{fig:maximprove}-left shows the testing result.

We also theoretically estimate the decoding robustness of our maximum likelihood
decoding method (\S\ref{sec:mld}).
Suppose the probability of correctly recognizing the glyph of a single letter is a constant $P_1$.
The probability $P_2$ of correctly decoding a single 5-letter block can be derived analytically:  
if we only use Chinese Remainder Decoding algorithm (\S\ref{sec:crc}),
$P_2$ is $\binom{5}{1}P_1^4(1-P_1)$. 
With the maximum likelihood decoding (\secref{sec:mld}), $P_2$ becomes $\binom{5}{2}P_1^3(1-P_1)^2$.
The improvement is visualized in \figref{fig:maximprove}-right.

\paragraph{Performance}
We use tensorflow\cite{tensorflow2015-whitepaper} with GPU support to train and decode the input.
It takes 0.89 seconds to decode a text sequence with 176 letters (30 blocks).
Our encoding algorithm is running in a single thread CPU, taking 7.28 seconds for the same length
of letters.
\endgroup

\begin{figure}[t]
 \centering
 \includegraphics[width=0.991\columnwidth]{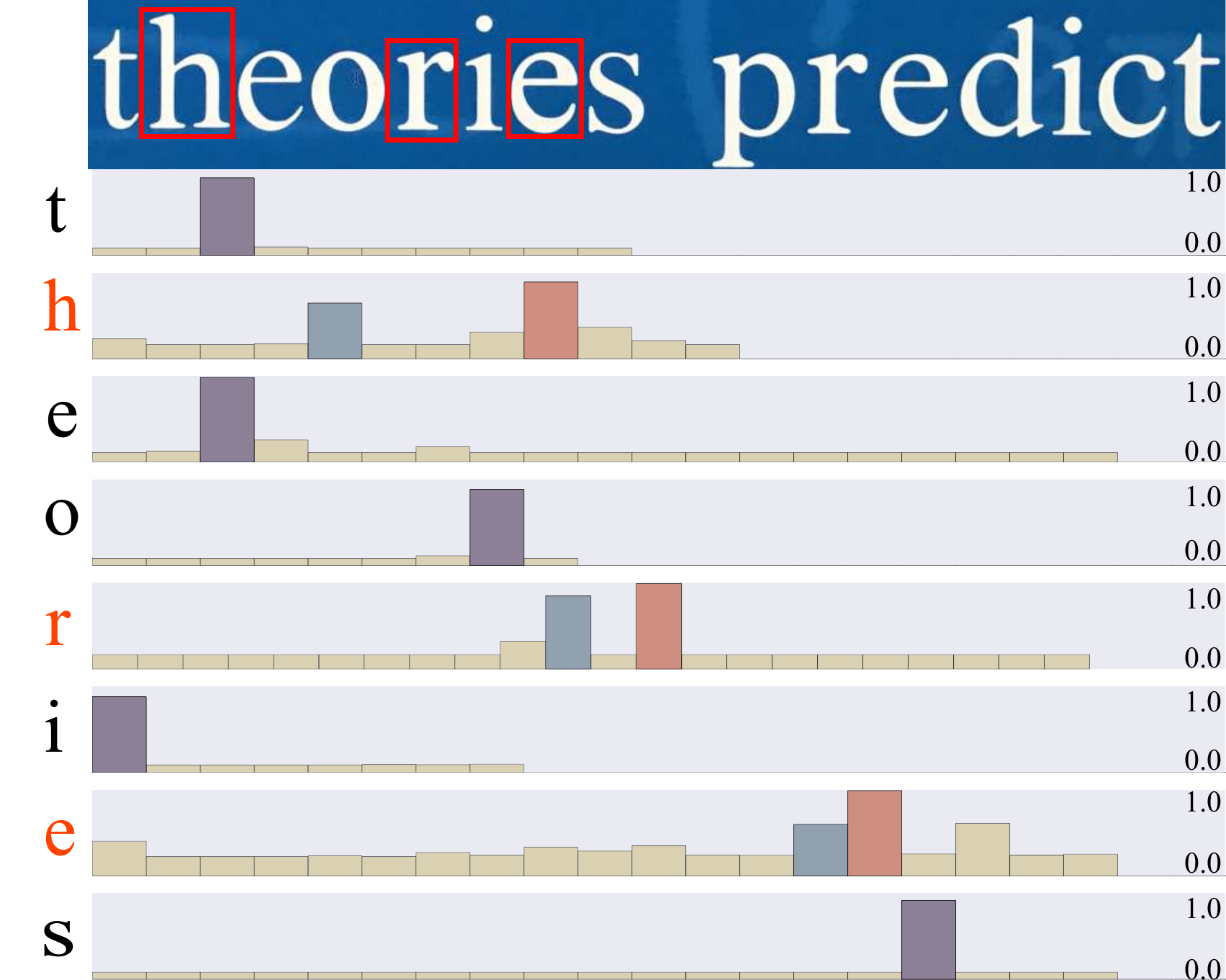}
 \vspace{-2.5mm}
 \caption{{\bf Decoding Probability.}
 {\bf (top)} A small region of photo to be decoded, where red boxes indicate recognition errors. 
 {\bf (bottom)} Each row visualizes the probabilities of recognizing input glyphs (from the
 image) as the perturbed glyphs. 
 Along the x-axis, every bar corresponds to a perturbed glyph in the codebook.
 The values on the y-axis are the output from our CNNs.  
 When an error occurs, the blue and red bars
 indicate the discrepancy between the correctly and incorrectly recognized glyphs. 
 \label{fig:distance}}
 \vspace{-4mm}
\end{figure}
\paragraph{Error correction improvement}
In \secref{sec:mld}, 
we hypothesize that the probability of recognizing an input pixel glyph $\bm{f}$ as a
glyph $\bm{u}$ in the codebook is proportional to the softmax output of the CNNs.
We validated this hypothesis experimentally.
Here we denote the softmax output value for recognizing $\bm{f}$ as a perturbed glyph $\bm{u}$ as $g(\bm{u}, \bm{f})$.
As shown in \figref{fig:distance} as an example, 
when $\bm{f}$ is mistakenly recognized as $\bm{u}$ as opposed to its true glyph $\bm{u}^*$,
the probability values $g(\bm{u},\bm{f})$ and $g(\bm{u}^*,\bm{f})$
are both high (although $d(\bm{u},\bm{f})$ is higher) and close to each other,
indicating that $\bm{f}$ may be recognized as $\bm{u}$ or $\bm{u}^*$ with close probabilities.
Thus, we conclude that using a likelihood model proportional to $g(\bm{u}^*,\bm{f})$ is reasonable.

\begin{figure}[t]
 \centering
 \includegraphics[width=0.90\columnwidth]{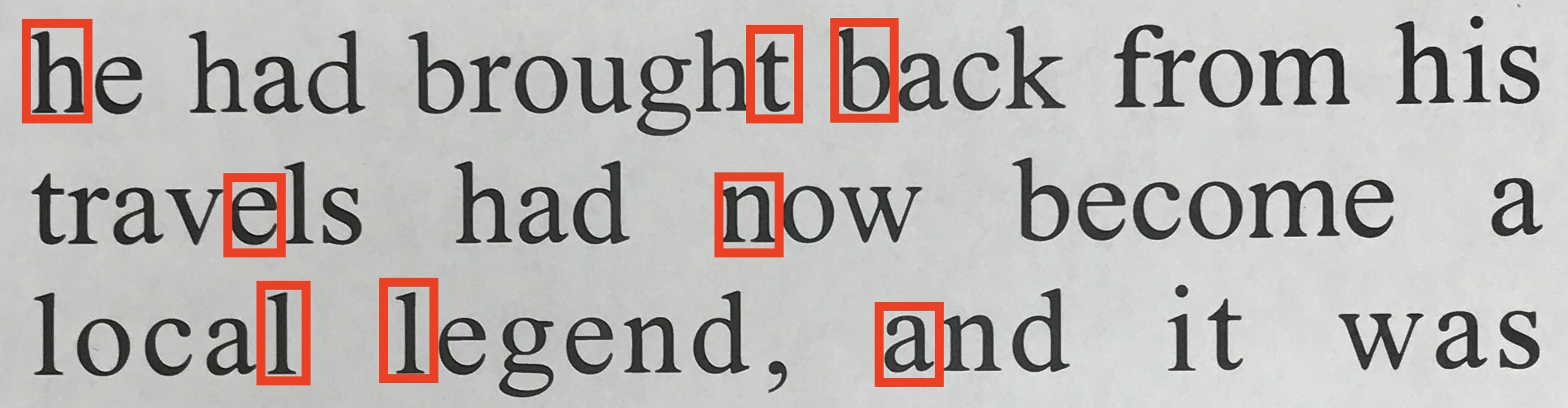}
 \vspace{-3mm}
 \caption{{\bf Correcting 2 errors.}
 We decode information from an input photo.
 Red boxes indicate where recognition errors occur.
 While there exist one block contains two errors, our decoding algorithm still
 successfully decodes.
 \label{fig:mld}}
 \vspace{-5mm}
\end{figure}

Additionally, we extensively tested our decoding algorithm on photos of text documents, and verified
that it can successfully decode all 5-letter blocks that have at most 2 errors.
A small snapshot of our tests is shown in \figref{fig:mld}.

\subsection{Perceptual Evaluation}\label{sec:valid}
To evaluate the subjective distortion introduced by perturbing the glyphs of a
document, we conducted two user studies on MTurk. Both studies follow the standard 2AFC protocol 
which is described in \S\ref{sec:perceptual} and has been used in other contexts. 

\begin{figure}[b]
  \vspace{-3mm}
  \centering
  \includegraphics[width=0.8\columnwidth]{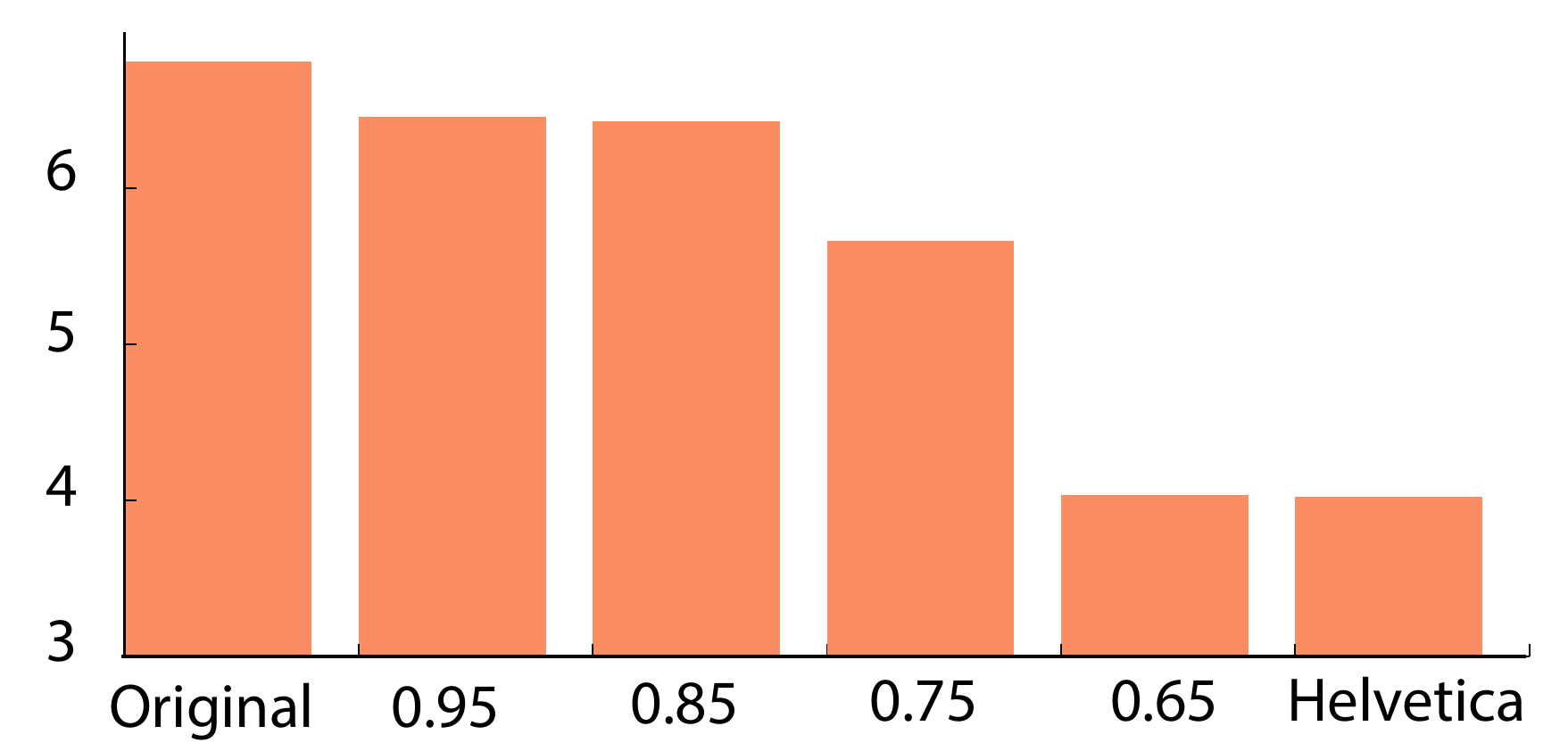}
  \vspace{-4mm}
  \caption{ {\bf User Study A.} 
  a to f are original font glyphs, 0.95, 0.85(we
  used), 0.75, 0.65, another different font, respectively. 
  }\label{fig:studyA}
\end{figure}
\emph{Study A} assesses the perceptual distortion of a perturbed glyph with respect to a standard font. 
We prepare a set of paragraphs, and the glyphs of each paragraph are
from one of the six categories: (1) the standard Times New Roman; (2-5) the perturbed glyphs from four glyph codebooks, 
in which the thresholds used to select the perceptually similar glyph
candidates are 0.95, 0.85 (i.e., the value we use in \S\ref{sec:perceptual}), 0.75, and 0.65, respectively;
and (6) a different font (Helvetica).
The font size of each paragraph ranges from 25pt to 60pt, so the number of letters in the paragraphs varies.
In each 2AFC question, we present the MTurk rater three short paragraphs: one
is in standard Times New Roman, the other two are randomly chosen from two of
the six categories. We ask the rater to select from the latter two paragraphs
the one whose font is closest to standard Times New Roman (shown in the
first paragraph).
We assign 16 questions of this type to each rater, and there were
169 raters participated. An example question is included in the supplemental document.  

After collecting the response, we use the same model~\eqref{eq:tb} to 
quantify the perceptual difference of the paragraphs in each of the six categories with respect to the 
one in standard Times New Roman. In this case, $s_i$ in \eqnref{eq:tb} is the perceptual difference
of a category of paragraphs to the paragraphs in standard Times New Roman. As shown in \figref{fig:studyA}, the 
results suggest that the glyphs in our codebook (generated with a threshold of 0.85)
lead to paragraphs that are perceptually close to the paragraphs in original glyphs---much closer than the glyphs 
selected by a lower threshold but almost as close as the glyphs selected by a higher threshold (i.e., 0.95).

\begin{figure}[t]
  \centering
  \includegraphics[width=0.90\columnwidth]{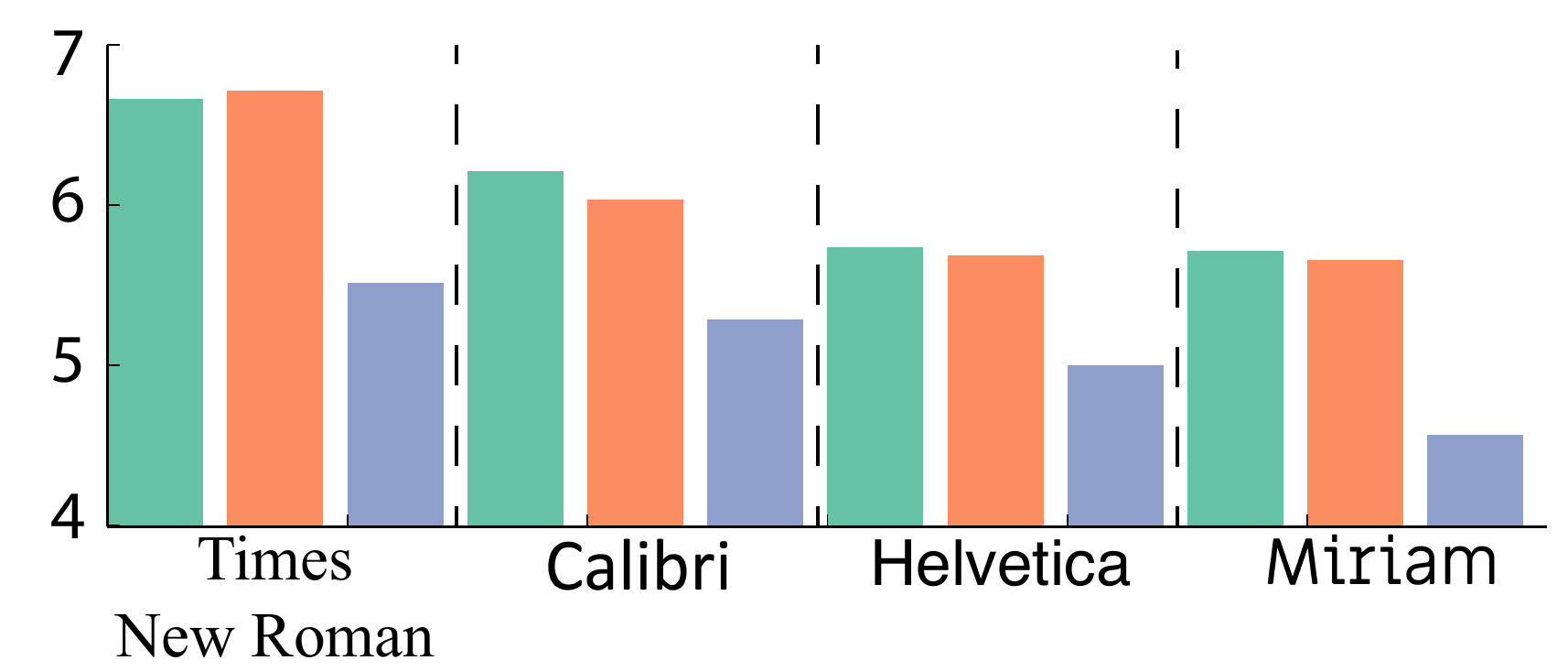}
  \vspace{-4mm}
  \caption{ {\bf User Study B.} 
  We conduct user studies for four different fonts simultaneously,
  and learn the scores of aesthetics when different types of glyphs are used:
  Green bars indicate the original glyphs, 
  orange bars indicate the perturbed glyphs from our codebook, 
  and purple bars indicate the perturbed glyphs chosen by a lower perceptual threshold (i.e., 0.7). } \label{fig:aesthetic}
  \figspace
\end{figure}

\emph{Study B} assesses how much the use of perturbed glyphs in a paragraph affects 
the aesthetics of its typeface. We prepare a set of paragraphs whose glyphs are from one of 
the 12 categories: We consider four different fonts (see \figref{fig:aesthetic}). For each font, we generate the glyphs in 
three ways, including (1) the unperturbed standard glyph; (2) the perturbed glyphs from our codebook using
a perceptual threshold of 0.85; and (3) the perturbed glyphs from a codebook using a threshold of 0.7.
In each 2AFC question, we present the MTurk rater two short paragraphs randomly chosen from two 
of the 12 categories, and ask the rater to select the paragraph whose typeface is aesthetically more pleasing to them.
We assign 16 questions of this type to each rater, and there were 135 participants. 
An example question is also included in the supplemental document.  

Again, using the logistic model~\eqref{eq:tb}, we quantify the aesthetics of the typeface
for paragraphs in the aforementioned three categories of glyphs.
\figref{fig:aesthetic} shows the results, indicating that,  
while the aesthetics are different across the four fonts,
for each font paragraphs using glyphs from our codebook (a threshold of 0.85)
are aesthetically comparable to the paragraphs in standard glyphs, 
whereas using glyphs selected by a lower perceptual threshold 
significantly depreciates the typeface aesthetics.

We note that in order to identify untrustworthy raters, in both user studies we include four control questions in each 
task assigned to the raters, in a way similar to those described in \S\ref{sec:perceptual}.

\section{Conclusion and Future Work}
We have introduced a new technique for embedding additional information in text documents.
Provided a user-specified message,
our method assigns each text letter an integer
and embeds the integer by perturbing the glyph of each letter according to a precomputed codebook.
Our method is able to correct a certain number of errors in the decoding stage,
through a new error-correction coding scheme built on two algorithmic
components: the Chinese Remainder coding and the maximum likelihood decoding.
We have demonstrated our method with four applications, 
including text document metadata storage, unobtrusive optical codes on text,
symmetric-key encryption, and format-independent digital signatures.


Currently we only consider standard fonts such as regular Times New Roman but
not their variants such as Times New Roman Bold Italic. We can treat those
variants separately and include their perturbed glyphs in the
codebook. Then, our method will work with those font variants.

When extracting messages from a rasterized text document, we rely on the OCR library
to detect and recognize the characters, but we cannot recover any OCR detection error.
If a character is mistakenly recognized by the OCR, the integer embedded in
that character is lost, and our error-correction decoding may not
be able to recover the plain message since different characters have different
embedding capacities.  However, in our experiments, the OCR library always recognizes characters correctly.

If a part of the text is completely occluded from the camera, our
decoding algorithm will fail, as it needs to know how the text is split
into blocks.  In the future, we hope to improve our coding scheme so that it is
able to recover from missing letters in the text as well. 

In general, the idea of perturbing the glyphs for embedding messages can be
applied to any symbolic system, such as other languages, mathematical
equations, and music notes. It would be interesting to explore 
similar embedding methods and their applications for different languages and symbols.

Lastly, while our method is robust to format conversion, rasterization, as well as photograph and
scan of printed papers, it suffers from the same drawback
that almost all text steganographic methods have: if a text document is completely
retyped, the embedded information is destroyed. 




\bibliographystyle{ACM-Reference-Format}
\bibliography{ref}

\SetKwComment{Comment}{$\triangleright$\ }{}
\begin{algorithm}[t]
    \SetKwInOut{Input}{Input}
    \SetKwInOut{Output}{Output}
    Initialize glyph candidates $\mathcal{C}$ \Comment*[r]{\S\ref{sec:perceptual}}
    $G \gets$ A complete graph whose nodes represent the elements of $\mathcal{C}$\;
    \While{true}{ 
        $Q\gets$Random select $M$ different pairs of nodes from $\mathcal{C}$\;
  \ForEach{pair $(\bm{u}_i,\bm{u}_j)$ in $Q$ }{

    $a \gets$ Training accuracy of CNN described using $(\bm{u}_i,\bm{u}_j)$ \Comment*[r]{\S\ref{subsec:codebook}}
    \If{$a<0.95$}{
      remove edge $(\bm{u}_i,\bm{u}_j)$ from $G$\;
    }
  }
  $G \gets$ Maximum Clique$(G)$\;
  $\mathcal{C} \gets$ glyphs corresponding to nodes in $G$\;
  \If{No change to $\mathcal{C}$}{break\;}
  }
  Train CNN using all elements in $\mathcal{C}$\;
  Remove elements whose recognition test accuracy is lower than 0.9\;
      return $\mathcal{C}$\;
   \caption{Confusion Test \label{alg:cftest}}
\end{algorithm}

\begin{algorithm}[t]
    \SetKwInOut{Input}{Input}
    \SetKwInOut{Output}{Output}
    \label{alg:decode}
    \Input{code vector $\bm{\bar{r}}=(\bar{r}_1,...\bar{r}_n)$, 
    mutually prime integers $\bm{p}=(p_1,...p_n)$, and $M$}
    \Output{Decoded integer $m$}
    $\tilde{m}\gets CRT(\bm{\bar{r}},\bm{p})$\;
    \eIf{$\tilde{m}<M$}{
    	$m \gets \tilde{m}$\;
        \textbf{return} success\;
    }{ Find codeword $\bm{\hat{r}}$, where $\bm{\hat{r}} = \min_{\bm{\hat{r}}} H(\boldsymbol{\hat{r}},\boldsymbol{\bar{r}})$\;
    	\eIf{$\boldsymbol{\hat{r}}$ is not unique}{
    		\textbf{return} fail\;
    	}
    	{
    		$m\gets CRT(\bm{\hat{r}},\bm{p})$\;
    		\textbf{return} success\;
    	}
    }

    \caption{Decoding}
\end{algorithm}

\end{document}